\documentclass[a4paper,fleqn]{cas-dc}

\usepackage[authoryear,longnamesfirst]{natbib}
\usepackage{microtype}
\usepackage{booktabs}      
\usepackage{tabularx}      
\usepackage{pifont}        
\usepackage[table]{xcolor} 
\usepackage{threeparttable}

\usepackage[ruled,vlined,linesnumbered]{algorithm2e}
\usepackage{stmaryrd} 
\def\tsc#1{\csdef{#1}{\textsc{\lowercase{#1}}\xspace}}
\tsc{WGM}
\tsc{QE}

\begin{document}
\let\WriteBookmarks\relax
\def\floatpagepagefraction{1}
\def\textpagefraction{.001}

\shorttitle{Federated Inference: Toward Privacy-Preserving Collaborative and Incentivized Model Serving}    

\shortauthors{Seo et al}  

\title [mode = title]{Federated Inference: Toward Privacy-Preserving Collaborative and Incentivized Model Serving}  



%

\author[1]{Jungwon Seo}[orcid=0000-0001-8174-9352]

\ead{jungwon.seo@uis.no}


\credit{Writing – original draft, Methodology, 
Writing – review \& editing}
\affiliation[1]{
            organization={Department of Electrical Engineering and Computer Science, University of Stavanger},
            city={Stavanger},
            postcode={4021}, 
            country={Norway}}
\affiliation[2]{
            organization={Department of Data Science, The Catholic University of Korea},
            city={Bucheon},
            postcode={14662}, 
            country={Republic of Korea}}

\author[1]{Ferhat Ozgur Catak}[orcid=0000-0002-2434-9966]
\ead{f.ozgur.catak@uis.no}
\credit{Writing – review \& editing}

\author[1]{Chunming Rong}[orcid=0000-0002-8347-0539]
\ead{chunming.rong@uis.no}
\credit{Writing – review \& editing, Funding acquisition}

\cortext[1]{Corresponding author}
\author[2]{Jaeyeon Jang}[orcid=0000-0001-6255-2044]
\ead{jaeyeon.jang@catholic.ac.kr}
\credit{Writing – review \& editing}
\cormark[1]

\begin{abstract}
Federated Inference (FI) studies how independently trained and privately owned models can collaborate at inference time without sharing data or model parameters.
While recent work has explored secure and distributed inference from disparate perspectives, a unified abstraction and system-level understanding of FI remain lacking. This paper positions FI as a distinct collaborative paradigm, complementary to federated learning, and identifies two fundamental requirements that govern its feasibility: inference-time privacy preservation and meaningful performance gains through collaboration. We formalize FI as a protected collaborative computation, analyze its core design dimensions, and examine the structural trade-offs that arise when privacy constraints, non-IID data, and limited observability are jointly imposed at inference time. Through a concrete instantiation and empirical analysis, we highlight recurring friction points in privacy-preserving inference, ensemble-based collaboration, and incentive alignment. Our findings suggest that FI exhibits system-level behaviors that cannot be directly inherited from training-time federation or classical ensemble methods. Overall, this work provides a unifying perspective on FI and outlines open challenges that must be addressed to enable practical, scalable, and privacy-preserving collaborative inference systems.
\end{abstract}


\maketitle
\section{Introduction}

Privacy-preserving machine learning has become a foundational requirement for modern data-driven systems deployed across organizational, legal, and geographical boundaries~\cite{li2019impact,chen2025privacy}.
To address concerns over data ownership and regulatory compliance, \emph{Federated Learning (FL)} has emerged as the dominant paradigm for collaborative model training, enabling multiple parties to jointly optimize a shared model while keeping raw data decentralized~\cite{mcmahan2017communication}.
As a result, FL has catalyzed extensive research across optimization, systems, and security~\cite{kairouz2021advances,mothukuri2021survey}, and is now widely regarded as the central abstraction for privacy-aware collaborative intelligence.

Despite its success, the assumptions underlying federated training often diverge from the realities of deployed AI ecosystems.
In many mature industrial and institutional settings, models are already pre-trained, legally isolated, and treated as proprietary strategic assets~\cite{philipp2020machine,RONG2025100277}.
Retraining such models via FL can be impractical—or even undesirable—due to operational costs, regulatory constraints, and intellectual property considerations~\cite{kuo2025research}.
In these environments, the primary opportunity for collaboration lies not in training, but in the inference stage.

Inference-stage collaboration introduces a fundamentally different interaction model.
Rather than exchanging gradients to update shared parameters, independently trained models can collaborate by jointly producing predictions for a given query.
This paradigm preserves full model ownership while enabling fine-grained, per-query interaction, thereby opening new possibilities for on-demand AI services and cross-organizational intelligence without centralized control~\cite{gan2023model,li2025collaborative}.

When privacy preservation is taken as a first-order requirement, inference-time collaboration must be considered under a different set of assumptions than those underlying generic collaborative or distributed inference.
In federated learning, the term ``federated'' does not simply denote distribution, but reflects a setting in which participating entities operate under limited mutual trust, constrained coordination, and restricted access to data and model parameters~\cite{zhang2022security}.
These constraints fundamentally shape system design choices and achievable performance.
The same perspective applies at the inference stage: privacy-preserving collaborative inference must operate under comparable limitations on control and information sharing.

We refer to inference-time collaboration under these constraints as \emph{Federated Inference (FI)}.
Although the term has appeared in a growing number of recent studies~\cite{fan2025federated,tanikanti2025first,alipanahloo2025rzkfl, zhu2024federated,jonker2025bayesian,vo2025federated,friston2024federated}, existing work typically addresses inference-time collaboration in a fragmented manner, focusing on individual technical aspects such as secure computation~\cite{zhao2025fedinf}, ensemble optimization~\cite{zhou2025towards}, or robustness~\cite{dhasade2025robust}.
As a result, FI is often treated as a setting-specific technique rather than as a coherent system-level problem.

A key observation motivating this work is that FI exposes a range of measurable factors beyond the question of whether collaboration is feasible.
Inference-time collaboration under limited trust and visibility introduces trade-offs among privacy protection, collaborative utility, and coordination constraints.
These trade-offs can be quantitatively evaluated and compared, suggesting that FI admits a structured design space rather than a binary notion of success or failure.

To study FI under these intertwined constraints, we introduce \emph{Federated Secure Ensemble Inference} (FedSEI).
FedSEI is a reference architecture that instantiates privacy-preserving collaborative inference in an end-to-end system, enabling controlled empirical analysis under realistic assumptions.
By integrating secure inference primitives~\cite{knott2021crypten}, ensemble-based collaboration, and a minimal coordination layer, FedSEI allows us to isolate and measure the system-level frictions that arise when privacy and collaboration are jointly enforced.

Through extensive experiments across diverse model architectures, data heterogeneity regimes, and network environments, we identify fundamental trade-offs inherent to FI.
Our results show that inference-time collaboration constitutes a distinct system design space with constraints that cannot be directly inherited from FL or conventional ensemble methods.
We view this work as a practical step toward enabling more systematic and cumulative research on FI grounded in system-level realities.

To summarize, our main contributions are:
\begin{itemize}
    \item \textbf{Federated Inference Design Space:}
    We study FI as a system-level collaboration problem and identify key axes along which inference-time collaboration can be quantitatively evaluated.

    \item \textbf{FedSEI Reference Architecture:}
    We design and implement \emph{FedSEI}, a complete reference architecture that enables privacy-preserving ensemble inference among independently trained models.

    \item \textbf{System-Level Empirical Analysis:}
    Using FedSEI, we empirically characterize the trade-offs and limitations that arise from enforcing privacy and collaboration at inference time.
\end{itemize}


\section{Related Work}
\subsection{The Evolution of Federated Learning Research}

FL has matured from a singular optimization problem into a multifaceted research ecosystem. The trajectory of FL research demonstrates how decentralized collaboration necessitates specialized solutions across distinct dimensions—spanning statistical optimization, system efficiency, security, and economics. This evolution serves as a blueprint for the emerging challenges in inference-stage collaboration.

Early FL research primarily focused on mitigating the impact of statistical (data) heterogeneity, which degrades model convergence. This led to a rich body of work on adaptive aggregation and personalized training strategies~\cite{reddi2021adaptive,oh2022fedbabu}.
Concurrently, to address the communication bottlenecks inherent in repeated model updates, the community developed extensive techniques for quantization, sparsification, and asynchronous coordination~\cite{haddadpour2021federated,nguyen2022federated}.
These efforts highlight that collaborative gain is inextricably linked to managing both statistical and computational frictions.

Beyond performance, FL has expanded to address the sociotechnical requirements of collaboration.
Given the risks of model poisoning and privacy leakage, a significant stream of research has integrated cryptographic primitives, differential privacy, and robust aggregation protocols to harden the training process~\cite{bonawitz2017practical,wei2020federated,kumar2023impact}.
Furthermore, recognizing that altruism is insufficient for sustained participation, recent works have formalized incentive mechanisms using game theory and blockchain technologies~\cite{ding2021incentive,RONG2025100277}.

The expansive landscape of FL research underscores a fundamental reality: decentralized AI is not merely an algorithmic challenge but a complex interplay of accuracy, latency, privacy, and incentives.
While FL addresses these dimensions during \emph{training}, our work posits that a similarly structured investigation is required for the \emph{inference} stage, where these constraints manifest differently.

\subsection{Evolution of Inference Architectures: From Cloud to Secure Collaboration}
\label{sec:inference_evolution}

The literature on model inference has evolved to address the competing requirements of performance, privacy, and domain adaptation. We classify existing approaches into centralized, local, and collaborative paradigms.

\paragraph{Centralized Cloud Inference.}
The dominant paradigm for model serving is Machine Learning as a Service (MLaaS), where high-capacity models are hosted on centralized clouds. While this approach maximizes computational efficiency and access to general-purpose models (e.g., LLMs), prior works have identified two critical limitations.
First, privacy risks: transmitting raw data to external servers violates data sovereignty requirements in regulated domains~\cite{shokri2017membership}.
Second, lack of domain specialization: generalist cloud models often underperform on specialized, proprietary datasets compared to models fine-tuned on local data.

\paragraph{Local Inference and the Utility Gap.}
To mitigate privacy risks, local inference serves as the natural alternative, where models are deployed directly on user devices or on-premise servers. This deployment strategy strictly maintains the data trust boundary.
However, this isolation introduces a "Utility Gap." Unlike centralized generalist models, local models operate in data silos, often suffering from the "Small Data" problem—exhibiting limited generalization due to the scarcity and bias of local training samples. To mitigate this, collaborative inference has been proposed to aggregate insights from multiple local models. Studies such as \emph{prediction ensembles}~\cite{yuan_mlink_2022,marsden2024universal} demonstrate that pooling predictions from diverse peer models can significantly improve robustness and accuracy without centralizing data.

\paragraph{Privacy-Preserving Primitives for Inference.}
One important challenge in collaborative inference is maintaining end-to-end confidentiality. The system must protect not only the model parameters but also the input queries and output predictions from being exposed to unauthorized parties. To address this, the literature has adapted privacy-preserving machine learning primitives:
\begin{itemize}
    \item \textbf{Homomorphic Encryption (HE):} Enables computation directly on encrypted data, ensuring that the server generates encrypted predictions without ever seeing the plaintext input or output~\cite{gentry2009fully}. However, it is often cited as computationally expensive for deep neural networks.
    \item \textbf{Trusted Execution Environments (TEEs):} Utilize hardware-isolated enclaves (e.g., SGX) to protect both the code (model) and data (input) from the host operating system~\cite{sagar2021confidential}. While efficient, they rely on hardware trust assumptions and are susceptible to side-channel attacks.
    \item \textbf{Secure Multi-Party Computation (SMPC):} Allows parties to jointly compute a function (e.g., aggregation) while keeping their respective inputs (user data) and weights (model parameters) private~\cite{knott2021crypten}. Recent works leverage SMPC to enable collaboration among mutually distrusting parties where no single entity can access the raw information.
\end{itemize}

Collectively, these primitives provide complementary mechanisms for enforcing confidentiality during inference, each reflecting different trade-offs between security assumptions and system efficiency.
They form the technical foundation upon which privacy-preserving collaborative inference systems can be constructed.

\subsection{Federated Inference Related Works}
\begin{table*}[t]
\caption{Detailed Taxonomy of Federated Inference Research. Recent literature utilizing the term "Federated Inference" spans across systems, statistics, trustworthy AI, and cognitive theory. This diversity highlights the need for a unified framework that integrates these technical dimensions within a unified system perspective.}
\label{tab:fi_taxonomy_final}
\centering
\resizebox{\textwidth}{!}{%
\begin{tabular}{@{}lll p{6cm} p{5.5cm} l@{}}
\toprule
\textbf{Domain} & \textbf{Reference} & \textbf{Year} & \textbf{Core Problem (The "Why")} & \textbf{Key Technique (The "How")} & \textbf{Target Setting} \\ \midrule
\multirow{4}{*}{\textbf{\begin{tabular}[c]{@{}l@{}}System \&\\ Efficiency\end{tabular}}} 
 & Zhao et al.~\cite{zhao2025fedinf} & 2025 & \textbf{Heterogeneity}: Overcoming convergence issues and high communication costs in standard FL. & \textbf{Hybrid Crypto}: Threshold HE for linear layers \& SMPC protocols (`SecReLU`/`Max`) for non-linear ones. & Cross-silo \\
 & Fan et al.~\cite{fan2025federated} & 2025 & \textbf{Resource Constraints}: Edge devices lack capacity for complex standalone inference. & \textbf{Workload Partitioning}: Collaborative task offloading and execution scheduling. & Edge Computing \\
 & Zhou et al.~\cite{zhou2025towards} & 2025 & \textbf{Concept Drift}: Retraining models for dynamic environments is computationally expensive. & \textbf{Online Ensemble}: Dynamic weight adjustment via contextual bandit algorithms. & Edge AI / IoT \\
 & Tanikanti et al.~\cite{tanikanti2025first} & 2025 & \textbf{Accessibility}: Difficulty in deploying and accessing scientific AI models on HPCs. & \textbf{Resource Scheduling}: A toolkit (FIRST) enabling "Inference-as-a-Service". & HPC Clusters \\ \midrule
\multirow{3}{*}{\textbf{\begin{tabular}[c]{@{}l@{}}Trustworthiness\\ (Security)\end{tabular}}} 
 & Dhasade et al.~\cite{dhasade2025robust} & 2025 & \textbf{Adversarial Attacks}: Malicious peers injecting poisoned predictions to corrupt results. & \textbf{Robust Aggregation}: Statistical outlier detection and defense mechanisms. & Adversarial \\
 & Alipanahloo et al.~\cite{alipanahloo2025rzkfl} & 2025 & \textbf{Verifiability}: Inability to verify the correctness of remote inference computation. & \textbf{Zero-Knowledge Proofs}: Recursive ZKPs to prove integrity without data leakage. & Decentralized \\
 & Zhu et al.~\cite{zhu2024federated} & 2024 & \textbf{Uncertainty}: Unreliable predictions due to noise and latency in wireless channels. & \textbf{Conformal Prediction}: Generating prediction sets with statistical coverage guarantees. & Wireless Net. \\ \midrule
\multirow{2}{*}{\textbf{\begin{tabular}[c]{@{}l@{}}Statistical \&\\ Causal\end{tabular}}} 
 & Jonker et al.~\cite{jonker2025bayesian} & 2025 & \textbf{Data Silos}: Inability to pool sensitive medical data for regression analysis. & \textbf{Bayesian Aggregation}: Aggregating local posterior parameters to approximate global inference. & Medical \\
 & Vo et al.~\cite{vo2025federated} & 2025 & \textbf{Hidden Confounders}: Distributed data obscures global causal structures. & \textbf{Structure Learning}: Algorithms to recover causal graphs from non-overlapping data. & Observational \\ \midrule
\textbf{Theory} 
 & Friston et al.~\cite{friston2024federated} & 2024 & \textbf{Cognitive Alignment}: How distributed intelligence emerges among agents sharing a common world. & \textbf{Active Inference}: Belief sharing and updating to minimize variational free energy. & Cognitive Science \\ \bottomrule
\end{tabular}%
}
\end{table*}

As summarized in Table~\ref{tab:fi_taxonomy_final}, recent studies using the term ``Federated Inference'' span a broad range of problem settings, including systems, statistics, security, and cognitive theory.
These works originate from different research communities and typically ground inference-time collaboration at distinct technical layers.

One line of work focuses on \emph{system efficiency}, addressing the feasibility of decentralized inference under practical constraints.
These studies examine whether collaborative inference can be executed at all in the presence of heterogeneous hardware, limited bandwidth, and privacy requirements.
Representative examples include cryptography-assisted inference pipelines in cross-silo settings~\cite{zhao2025fedinf}, workload partitioning for edge devices~\cite{fan2025federated}, and resource scheduling frameworks for large-scale infrastructures~\cite{tanikanti2025first}.
Together, these efforts demonstrate that inference-time collaboration is operationally viable across diverse deployment environments.

A second line of work emphasizes the \emph{quality and reliability} of collaborative inference.
Here, FI is often framed as an aggregation or ensemble problem, with the goal of improving predictive performance or robustness relative to isolated inference.
Examples include online ensemble optimization under concept drift~\cite{zhou2025towards}, defenses against adversarial participants~\cite{dhasade2025robust}, uncertainty-aware prediction in noisy wireless settings~\cite{zhu2024federated}, and verifiable inference using zero-knowledge proofs~\cite{alipanahloo2025rzkfl}.

Taken together, these studies highlight multiple dimensions along which FI systems can be evaluated, including feasibility, utility, and trustworthiness.
However, these dimensions are typically examined in isolation and under differing assumptions regarding system ownership, participant behavior, and deployment context.
As a result, existing work provides valuable but fragmented insights into inference-time collaboration, making it challenging to reason about FI as a unified system problem.

From this perspective, prior approaches can be viewed as addressing complementary aspects of FI at different technical layers.
This work builds on these efforts by adopting a system-level viewpoint that connects these dimensions and enables more structured and cumulative investigation of FI.


\section{Federated Inference}
\label{sec:fi}

FI studies inference-time collaboration among multiple models that are independently owned, operated, and protected by different parties.
Unlike FL, where collaboration is achieved through repeated model updates, FI focuses on producing a prediction for a specific query while preserving model autonomy and confidentiality. Table~\ref{tab:notation} summarizes the key notations used throughout the formulation.

\begin{table}[t]
\centering
\caption{Summary of Key Notations}
\label{tab:notation}
\resizebox{\columnwidth}{!}{%
\begin{tabular}{l l}
\toprule
\textbf{Symbol} & \textbf{Description} \\
\midrule
$N$ & Number of model owners \\
$P_i$ & The $i$-th model owner \\
$M_i$ & Model owned by provider $P_i$ \\
$C$ & Client requesting inference \\
$x$ & Private input data ($x \in \mathcal{X}$) \\
$y$ & Final collaborative inference result \\
$\llbracket \cdot \rrbracket$ & Value protected by a privacy-preserving mechanism \\
$\mathcal{G}$ & Abstract collaboration function \\
$y_i$ & Intermediate (protected) prediction from model $M_i$ \\
$w_i$ & Aggregation weight assigned to model $M_i$ \\
\bottomrule
\end{tabular}%
}
\end{table}

\subsection{Problem Setting}

\paragraph{Federated Inference at a Glance.}
At a high level, an FI system consists of three conceptual roles:
\begin{itemize}
    \item A \textbf{client} $C$ holding a private input $x \in \mathcal{X}$ and requesting inference.
    \item A set of \textbf{model owners} $\{P_1, \dots, P_N\}$, each owning an independently trained and proprietary model $M_i$.
    \item An \textbf{execution environment} that enables inference-time collaboration while enforcing privacy constraints.
\end{itemize}

Collaboration in FI occurs exclusively at inference time.
Models are static: they are not updated, synchronized, or retrained as part of the inference process.
Each inference request is handled independently, and collaboration is limited to producing a prediction for a given query.

\paragraph{Privacy as a Defining Constraint.}
Privacy preservation is not an optional feature of FI but a defining constraint.
An inference system is considered federated only if it satisfies the following confidentiality requirements:
\begin{itemize}
    \item \textbf{Input confidentiality}: the client input $x$ is not revealed in plaintext.
    \item \textbf{Model confidentiality}: the parameters of each model $M_i$ remain private to its owner.
    \item \textbf{Intermediate confidentiality}: intermediate representations and per-model predictions are not observable in plaintext.
\end{itemize}

The final output $y$ may be revealed to the client or to a restricted set of parties, depending on the application.
Any approach that relies on direct access to plaintext inputs, model parameters or internal representations, or per-model intermediate predictions lies outside
the scope of FI.

\paragraph{Threat Model and Trust Assumptions.}
FI operates under limited trust assumptions that reflect realistic cross-silo deployments:
\begin{itemize}
    \item \textbf{Model owners are mutually distrustful} and should not learn proprietary information about other models.
    \item \textbf{The execution environment is not trusted} with plaintext access to inputs, models, or intermediate values.
    \item Participating entities may range from \textbf{honest-but-curious} to \textbf{malicious}, depending on the application and deployment context.
\end{itemize}

We assume that collusion is bounded such that no single entity gains visibility into all sensitive assets in plaintext.
These assumptions define the adversarial surface considered in FI and act as hard constraints on admissible system designs.

\subsection{Problem Formulation}

We now formalize FI as a protected collaborative computation.

\paragraph{Protected Inference Computation.}
Under the above constraints, FI can be abstracted as a protected computation.
Let $\llbracket \cdot \rrbracket$ denote a value protected by a privacy-preserving mechanism.
The inference computation is expressed as
\begin{equation}
y = \mathcal{G}\big(\llbracket x \rrbracket, \{\llbracket M_i \rrbracket\}_{i=1}^{N}\big),
\end{equation}
where $\mathcal{G}$ represents a collaboration function.

The function $\mathcal{G}$ encapsulates model evaluation and collaboration logic—such as aggregation, routing, or fusion—while ensuring that all intermediate values remain protected throughout execution.
Plaintext reconstruction occurs only at the final output stage.

\paragraph{Design Axes within the FI Constraint Space.}
While the confidentiality requirements above define the boundary of FI, system behavior within this boundary is governed by a set of design axes:
\begin{itemize}
    \item \textbf{Privacy strength}: which entities are considered adversarial and what forms of information leakage are tolerated.
    \item \textbf{Collaborative utility}: how model contributions are combined and the resulting performance gain over isolated inference.
    \item \textbf{System efficiency}: latency, communication overhead, and scalability with respect to the number of participating models.
\end{itemize}

These axes admit quantitative evaluation and trade-offs.
Consequently, FI defines a structured design space rather than a binary notion of feasibility.

\section{Instantiation of Federated Inference}
\label{sec:inst-FedSEI}

\begin{figure*}[t]

    \centering

    \includegraphics[width=0.9\linewidth]{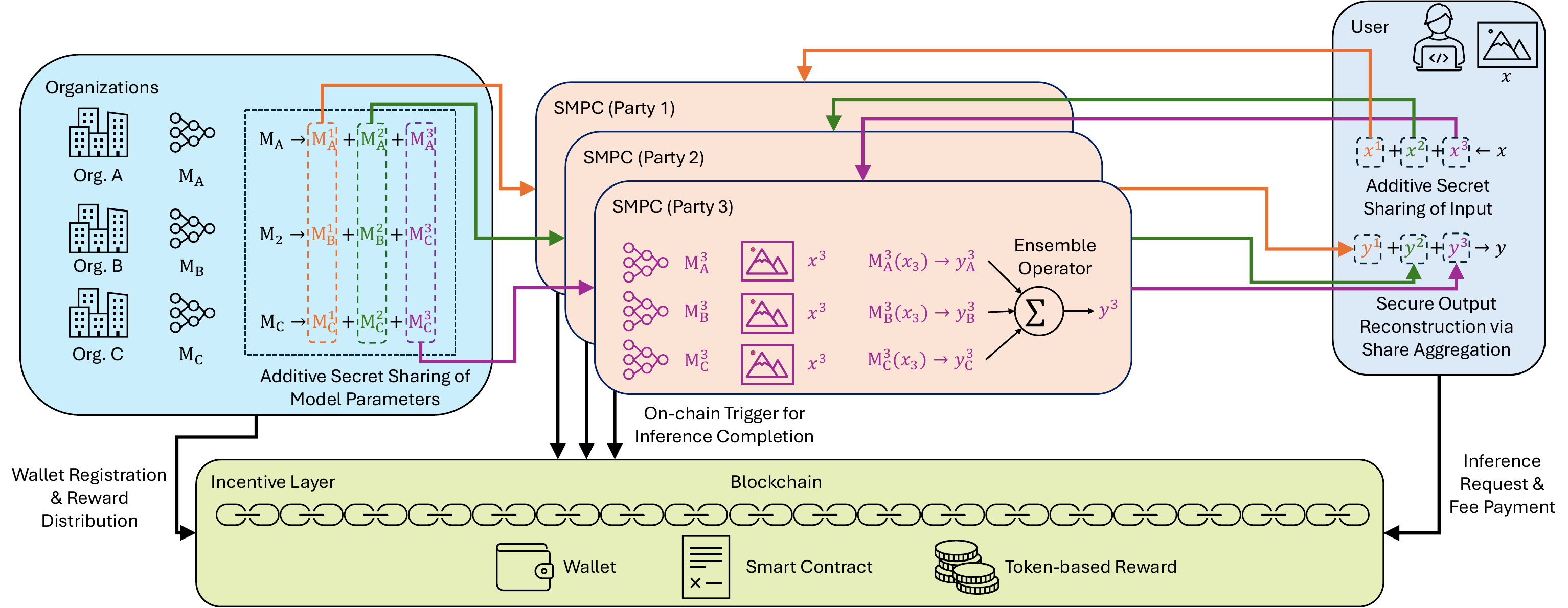}

    \caption{
System architecture and workflow of FedSEI.
Models owned by different organizations are additively secret-shared across multiple SMPC parties, which jointly perform privacy-preserving inference and ensemble aggregation.
Lower-script indices (e.g., $M_A$) denote model owners, while upper-script indices (e.g., $M_A^k$, $x^k$, $y^k$) denote additive secret shares of models, inputs, and outputs held by SMPC party $k$.
The abstract protected values $\llbracket x \rrbracket$, $\llbracket M_i \rrbracket$, and $\llbracket y \rrbracket$ used in the main text correspond to the collection of these shares.
}

    \label{fig:system-overview}

\end{figure*}

We now introduce \emph{FedSEI}, a concrete instantiation of the FI framework. Inference-time privacy is ensured via SMPC, which guarantees that both model parameters and intermediate representations remain protected throughout execution.
Collaboration is achieved by aggregating the protected outputs of multiple independently owned models, enabling ensemble-based prediction without revealing individual model outputs.

In addition, FedSEI incorporates smart contracts as an auxiliary system component to support incentive mechanisms and auditable collaboration.
While incentive design is not treated as a defining requirement of FI, this integration demonstrates how practical deployment considerations can be layered on top of the core FI abstraction.

\subsection{Preliminaries: SMPC and Secret Sharing} 

To address the privacy constraints, FedSEI relies on SMPC based on \emph{Additive Secret Sharing}.
In this setting, a private value $x$ is split into random shares $\{x_1, \dots, x_N\}$ such that $\sum x_i \equiv x \pmod q$.
Each share $x_k$ is distributed to a distinct computing server (or \emph{party}).
Crucially, this scheme satisfies the closure property for arithmetic operations:
computing on shares (e.g., $\llbracket x \rrbracket + \llbracket y \rrbracket$ or $\llbracket x \rrbracket \cdot \llbracket y \rrbracket$) yields a valid share of the result (e.g., $\llbracket x+y \rrbracket$) without ever revealing the underlying plaintext.
We adopt CrypTen~\cite{knott2021crypten}, a PyTorch-compatible and general-purpose SMPC framework, which allows us to implement and evaluate a range of ensemble aggregation strategies entirely within the secure computation domain.

\subsection{Collaborative Inference via SMPC}

To support a broad range of collaborative inference strategies,
we formulate the inference stage in FedSEI as a generalized
\emph{secure aggregation} process executed entirely over protected values.

\paragraph{Protected Model Inference.}
Given a protected input $\llbracket x \rrbracket$,
each participating model—now represented as a protected object $\llbracket M_i \rrbracket$—performs
a protected forward pass. This operation produces an encrypted prediction:
\begin{equation}
\llbracket y_i \rrbracket = \llbracket M_i \rrbracket(\llbracket x \rrbracket),
\end{equation}
where $\llbracket M_i \rrbracket(\cdot)$ denotes the evaluation of the secret-shared model
on the secret-shared input.
Mathematically, this implies that all linear (e.g., matrix multiplications) and non-linear
operations within the model are executed using SMPC protocols corresponding to the shares held by the parties.
Depending on the task, $\llbracket y_i \rrbracket$ may correspond to logits, probability vectors, or regression scores.

\paragraph{Generalized Secure Aggregation.}
The individual protected predictions $\{\llbracket y_i \rrbracket\}_{i=1}^{N}$
are combined through a secure aggregation function $\mathcal{A}$,
which operates entirely in the protected domain:
\begin{equation}
\llbracket y \rrbracket
= \mathcal{A}\big( \{ w_i, \llbracket y_i \rrbracket \}_{i=1}^{N} \big).
\end{equation}
In our reference implementation, we instantiate $\mathcal{A}$ as a weighted aggregation:
\begin{equation}
\llbracket y \rrbracket
= \sum_{i=1}^{N} w_i \cdot \llbracket y_i \rrbracket,
\qquad \text{s.t.} \quad \sum_{i=1}^{N} w_i = 1.
\end{equation}

\subsection{System Architecture}
Figure~\ref{fig:system-overview} illustrates the end-to-end architecture of FedSEI.
The system involves three distinct types of entities, clarifying the separation between model ownership and computation:
\begin{enumerate}
    \item \textbf{Model Organizations (Providers):} Entities that own proprietary pre-trained models ($M_i$).
    \item \textbf{Client (User):} An end-user who holds private input data ($x$) and requests inference.
    \item \textbf{Computing Parties ($\mathbf{P}$):} A set of semi-honest servers that execute the SMPC protocol. To prevent collusion, the Client or a Model Provider can participate as a computing node. 
\end{enumerate}
We implement FedSEI using \texttt{FastAPI}\footnote{\url{https://fastapi.tiangolo.com/}} as a lightweight communication interface deployed independently at each SMPC party.
Each party exposes a FastAPI endpoint that wraps the underlying CrypTen-based MPC execution engine.
To avoid reliance on any intermediate coordinator, the client directly communicates with all participating parties for input submission and output reconstruction, ensuring that inference-time interaction remains fully peer-to-peer. System-wide functionalities such as party registration, participation management, and cost settlement are handled externally via smart contracts, decoupling orchestration and economic logic from the inference pipeline. This design results in a fully decentralized architecture, where no single entity assumes a trusted coordinating role during secure inference.
Algorithm~\ref{alg:FedSEI} summarizes the complete workflow.
To support reproducibility, we provide an open-source prototype implementation of FedSEI at \url{https://github.com/thejungwon/FedSEI}.

\paragraph{Security Assumptions and Threat Model.}
FedSEI assumes a semi-honest (honest-but-curious) adversary model, in which all participating entities correctly follow the prescribed protocols but may attempt to infer additional information from observed protocol transcripts.

We consider a cross-silo deployment setting, where each participating organization both provides a model and directly participates as a computing party in the SMPC protocol.
Accordingly, we assume a one-to-one correspondence between model providers and computing parties.
Inference is executed via additive secret sharing among a set of computing parties $\mathbf{P}$.

Under the semi-honest adversary model, additive secret sharing provides confidentiality against collusion of up to $t = |\mathbf{P}| - 1$ computing parties, as long as at least one party remains honest.
This design ensures that a model provider’s private information cannot be compromised without its own participation in collusion.
In contrast to settings where computation is delegated to external operators, model providers in FedSEI retain direct control over the confidentiality of their models, and information leakage is solely bounded by the SMPC collusion threshold.

Inference inputs are assumed to originate from one of the participating silos and are therefore protected under the same trust assumptions as model parameters.
More general deployment scenarios involving arbitrary external clients are not considered in this work.
While similar input-level confidentiality guarantees could be achieved by allowing clients to dynamically register as computing parties, such mechanisms are beyond the scope of this paper.

We do not consider active or malicious deviations, integrity attacks on computation, or adversarial manipulation of model parameters.
The on-chain incentive mechanism verifies protocol completion via authenticated signatures from the computing parties, but does not provide cryptographic guarantees regarding semantic correctness, model identity, or weight provenance.

\subsection{Workflow Overview}
The workflow proceeds through the following phases.

\paragraph{1. Secure Model Provisioning.}
To participate without revealing raw parameters, Model providers apply additive secret sharing to their model parameters.
As illustrated in Figure~\ref{fig:system-overview} (left), a model $M_i$ is effectively converted into a protected object $\llbracket M_i \rrbracket$.
Its additive secret shares (e.g., $M_i^1, M_i^2, M_i^3$) are distributed across the distinct SMPC parties.
This ensures that no single party holds the complete model.

\paragraph{2. Privacy-Preserving Inference.}
When a Client requests inference, they secret-share their input $x$ and directly send the shares to the SMPC Parties.
As depicted in the center of Figure~\ref{fig:system-overview}, the parties jointly execute the inference logic $\llbracket y_i \rrbracket = \llbracket M_i \rrbracket(\llbracket x \rrbracket)$ and the ensemble aggregation entirely in the encrypted domain.
Due to the closure property of SMPC, the intermediate results remain as secret shares throughout the computation.

\paragraph{3. Result Reconstruction and Incentive.}
After computation, the output shares are returned to the Client for local reconstruction.
In parallel with this workflow (Figure~\ref{fig:system-overview}, right and bottom), the system enforces an on-chain incentive mechanism.
Using Ethereum smart contracts, the system enforces a "commit-then-reveal" workflow: the client deposits a fee before execution, and the contract releases rewards to model providers only after the SMPC parties provide cryptographic proof (signatures) of job completion.

\begin{algorithm}[t]
\small
\caption{FedSEI Workflow}
\label{alg:FedSEI}
\KwIn{
    Private input $x$ from client $C$,
    Protected models $\{\llbracket M_i \rrbracket\}_{i=1}^{N}$,
    Computing parties $\mathbf{P} = \{P_1, \dots, P_K\}$,
    Job identifier $j$.
}
\KwOut{Final prediction $y$}

\BlankLine
\textbf{Phase 1: Input Secret Sharing}\;
Client generates protected input:
$\llbracket x \rrbracket \leftarrow \textsc{Share}(x)$\;
Client sends share $\llbracket x \rrbracket^{k}$ to each computing party $P_k \in \mathbf{P}$\;

\BlankLine
\textbf{Phase 2: On-chain Escrow Initialization}\;
Client $C$ invokes \textsc{CreateJob}$(j, \mathbf{P})$ with deposit $d$\;
Computing parties verify that job $j$ is registered and active on-chain\;

\BlankLine
\textbf{Phase 3: Privacy-Preserving Model Execution}\;
\ForEach{model $i \in \{1, \dots, N\}$}{
    Parties jointly compute the protected prediction:\;
    \Indp
    $\llbracket y_i \rrbracket \leftarrow \llbracket M_i \rrbracket(\llbracket x \rrbracket)$\;
    \Indm
}

\BlankLine
\textbf{Phase 4: Secure Ensemble Aggregation}\;
\ForEach{party $P_k \in \mathbf{P}$}{
    Compute local output share:\;
    \Indp
    $
    \llbracket y \rrbracket^{k}
    \leftarrow
    \sum_{i=1}^{N} w_i \cdot \llbracket y_i \rrbracket^{k}
    $\;
    \Indm
}

\BlankLine
\textbf{Phase 5: Result Reconstruction and Settlement}\;
    \tcc{$H(\cdot)$ is a cryptographic hash function; $sk_k$ is the signing key of party $P_k$}

\ForEach{party $P_k \in \mathbf{P}$}{
    Send result share $\llbracket y \rrbracket^{k}$ to client $C$\;
    Generate proof $\sigma_k \leftarrow \textsc{Sign}_{sk_k}(H(j \parallel C))$\;
    Send $\sigma_k$ to client $C$\;
}
Client reconstructs output:\;
$
y \leftarrow
\textsc{Reconstruct}(\{\llbracket y \rrbracket^{k}\}_{k=1}^{K})
$\;

Client invokes \textsc{CompleteJob}$(j, \{\sigma_k\})$ to release rewards\;

\Return{$y$}
\end{algorithm}

\begin{algorithm}[t]
\small
\caption{Escrow Smart Contract}
\label{alg:sc_escrow}
\KwIn{
    Job ID $j$,
    Set of parties $\mathbf{P} = \{P_1, \dots, P_N\}$,
    Deposit $d$,
    Set of signatures $\Sigma = \{\sigma_1, \dots, \sigma_N\}$
}
\KwOut{Reward transfers to parties}

\BlankLine
\textbf{CreateJob}$(j, \mathbf{P})$:\\
\If{$jobs[j]$ exists \textbf{or} $d = 0$}{
    \Return reject\;
}
Store $jobs[j] \leftarrow (\text{client}=C,\ \text{parties}=\mathbf{P},\ \text{deposit}=d,\ \text{completed}=\texttt{false})$\;

\BlankLine

\textbf{CompleteJob}$(j, \Sigma)$:\\
\tcc{Job validity and client authorization checks}
Load $job \leftarrow jobs[j]$\;
\If{$job$ is missing \textbf{or} $job.completed = \texttt{true}$}{
    \Return reject\;
}
\If{caller $\neq job.client$}{
    \Return reject\;
}

\BlankLine
\tcc{Verify signatures from all registered parties}
$m \leftarrow H(j \parallel job.client)$\;
\ForEach{party $P_k \in job.parties$}{
    \If{\textsc{Verify}$(m, \sigma_k, P_k.\text{pubKey})$ is \texttt{false}}{
        \Return reject\;
    }
}

\BlankLine
\tcc{Distribute rewards via policy function}
$\{r_k\}_{P_k \in \mathbf{P}} \leftarrow \textsc{AllocateReward}(job.deposit, \mathbf{P})$\;
\ForEach{party $P_k \in \mathbf{P}$}{
    Transfer amount $r_k$ to address $P_k$\;
}
Set $job.completed \leftarrow \texttt{true}$\;
\end{algorithm}

\subsection{Incentive Mechanism}
FedSEI incorporates an on-chain incentive mechanism to enable decentralized and accountable settlement for collaborative inference.
Using Ethereum smart contracts~\cite{buterin2014next}, the system eliminates the need for a central authority by enforcing automatic service settlement based on authenticated participation of the computing parties.
We adopt an Ethereum-based implementation as it requires no additional infrastructure setup and provides immediate deployability using existing public blockchain ecosystems.
At the same time, the incentive layer is designed to be modular: in scenarios where transaction fees or on-chain information exposure are of concern, the same logic can be realized using permissioned blockchain frameworks such as Hyperledger~\cite{androulaki2018hyperledger}, without affecting the core inference pipeline.

The incentive workflow follows Algorithm~\ref{alg:sc_escrow}.
A client first deposits a service fee into the contract.
Upon completion of inference, the computing parties submit cryptographic signatures $\{\sigma_k\}_{P_k \in \mathbf{P}}$ attesting to successful protocol execution (e.g., agreement on the final output hash).
Under the assumed semi-honest setting with $(N,N)$-secret sharing, the system operates under a fail-stop model: if any party aborts or withholds a signature, the protocol terminates without revealing the result in order to preserve privacy.
The smart contract verifies the submitted signatures against the registered public keys and releases the escrowed payment only upon successful verification.

In the current implementation, rewards are distributed uniformly among participating computing parties, i.e., each party receives $1/|\mathbf{P}|$ of the service fee.
This choice reflects the fundamental difficulty of objectively measuring individual contribution in collaborative inference settings without access to ground-truth labels or trusted performance oracles.
While more sophisticated reward allocation mechanisms are desirable, designing contribution-aware and manipulation-resistant reward functions remains a challenging open problem.
We defer a detailed discussion of reward fairness and its inherent limitations to Section~\ref{subsec:reward_fairness}.

\section{Experiments and Analysis}
\label{sec:experiments}

The objective of this section is to empirically characterize the practical friction points that arise across the functional stages of FI.
Rather than reporting a single end-to-end metric, we conduct a component-wise evaluation to expose how privacy enforcement, collaboration under heterogeneity, and incentive mechanisms each affect system behavior.
This design enables a precise examination of when inference-stage collaboration is operationally viable, performance-wise beneficial, and economically sustainable.

Accordingly, we structure our experimental analysis along three complementary axes that correspond to the core technical requirements of collaborative inference:

\begin{itemize}
    \item \textbf{Inference-time Privacy Overhead:}
    We quantify the computational and communication costs introduced by privacy-preserving inference protocols as a function of model complexity and the number of participating parties.

    \item \textbf{Collaborative Performance under Heterogeneity:}
    We evaluate the robustness of ensemble-based collaboration under model and data heterogeneity, focusing on predictive behavior and convergence properties of the collaborative logic.

    \item \textbf{Incentive Alignment and Sustainability:}
    We analyze reward allocation behavior in environments without ground-truth labels to assess the feasibility and limitations of incentive mechanisms for inference-stage collaboration.
\end{itemize}

Through this structured evaluation, we address the following research question:
\textit{Under what systemic constraints can collaborative inference be realized in a manner that is both technically feasible and economically meaningful?}

\subsection{SMPC Computational Overhead}\label{sec:computation_overhead}

\begin{table*}[htbp]
\centering
\renewcommand{\arraystretch}{1.3}
\caption{Operational Cost Analysis of Neural Layers under Secret-Sharing MPC (CrypTen). 
\textbf{Round Complexity} refers to the number of sequential communication steps, required by interactive MPC protocols. 
\textbf{Notations:} $D$ denotes the input/output feature dimension of a linear layer, with bandwidth cost scaling as $O(D^2)$, 
$\ell$ is the bit-length for secure comparisons, and $d$ is the polynomial degree for approximations.}
\scriptsize
\label{tab:mpc_ops_refined}
\resizebox{\linewidth}{!}{
\begin{tabular}{lcccc}
\toprule
\textbf{Operation} & \textbf{MPC Protocol} & \textbf{Round Complexity} & \textbf{Bottleneck Severity} & \textbf{Primary Factor} \\
\midrule
Addition / Scalar Mult. & Local Sharing & $0$ & Low & No Communication \\
Matrix Mult. (Linear) & Beaver Triples & $O(1)$ & \textbf{High} & Bandwidth Volume ($O(D^2)$) \\
ReLU / MaxPool & Secure Comparison & $O(\ell)$ & \textbf{Very High} & Latency (Bit-level Round Trips) \\
Softmax / Sigmoid & Poly/Chebychev Approx. & $O(d)$ & \textbf{Critical} & Multi-round Interactivity \\
BatchNorm (Inference) & Affine Transform & $0$ & Low & Pre-computed Stats \\
\bottomrule
\end{tabular}
}
\end{table*}

\paragraph{Motivation and Scope.}
This subsection examines the computational overhead of SMPC-based neural network inference with network communication excluded.
Table~\ref{tab:mpc_ops_refined} summarizes the operational characteristics of core neural network layers under a CrypTen-based secret-sharing MPC implementation, including their round complexity and dominant cost factors.
These costs reflect the intrinsic computational and synchronization overhead of MPC primitives, separate from additional delays introduced by network communication. All experiments in this subsection are conducted on a single host. All MPC parties are executed on the same physical machine using local multiprocessing, which removes network latency while preserving protocol-level synchronization and execution semantics.
The impact of network communication on end-to-end inference latency is analyzed separately in Section~\ref{sec:network_overhead}.

\paragraph{Model Architectures.}
We consider a spectrum of neural network complexities to assess scalability, ranging from lightweight multi-layer perceptrons (MLPs) to widely adopted convolutional architectures.
Specifically, we evaluate MLPs with varying capacities and standard convolutional neural networks (CNNs), whose architectural details are summarized in Table~\ref{tab:models}.
In addition, we include benchmark models such as LeNet~\cite{lecun2002gradient} and ResNet-18~\cite{he2016deep}, which serve as baselines for comparison across heterogeneous system configurations.

\begin{table}[h]
\centering
\caption{Model architectures used in our experiments (ordered by execution in \texttt{forward}).}
\label{tab:models}
\resizebox{\linewidth}{!}{
\begin{tabular}{lcccccc}
\toprule
\textbf{Step} & \textbf{SmallMLP} & \textbf{MediumMLP} & \textbf{LargeMLP} &
\textbf{SmallCNN} & \textbf{MediumCNN} & \textbf{LargeCNN} \\
\midrule
1 &
FC(3072$\rightarrow$256) &
FC(3072$\rightarrow$512) &
FC(3072$\rightarrow$1024) &
Conv(3$\rightarrow$32) &
Conv(3$\rightarrow$32) &
Conv(3$\rightarrow$64) \\
2 &
FC(256$\rightarrow$10) &
FC(512$\rightarrow$256) &
FC(1024$\rightarrow$512) &
MaxPool(2$\times$2) &
MaxPool(2$\times$2) &
Conv(64$\rightarrow$64) \\
3 &
-- &
FC(256$\rightarrow$10) &
FC(512$\rightarrow$256) &
Conv(32$\rightarrow$64) &
Conv(32$\rightarrow$64) &
MaxPool(2$\times$2) \\
4 &
-- &
-- &
FC(256$\rightarrow$10) &
MaxPool(2$\times$2) &
Conv(64$\rightarrow$128) &
Conv(64$\rightarrow$128) \\
5 &
-- &
-- &
-- &
FC(4096$\rightarrow$256) &
MaxPool(2$\times$2) &
Conv(128$\rightarrow$128) \\
6 &
-- &
-- &
-- &
FC(256$\rightarrow$10) &
FC(8192$\rightarrow$512) &
MaxPool(2$\times$2) \\
7 &
-- &
-- &
-- &
-- &
FC(512$\rightarrow$10) &
FC(8192$\rightarrow$512) \\
8 &
-- &
-- &
-- &
-- &
-- &
FC(512$\rightarrow$256) \\
9 &
-- &
-- &
-- &
-- &
-- &
FC(256$\rightarrow$10) \\
\midrule
\textbf{\#Params} &
789,258 &
1,707,274 &
3,805,450 &
1,070,794 &
4,293,194 &
4,588,874 \\
\bottomrule
\end{tabular}
}
\end{table}

\paragraph{Experimental Setup.}
We measure inference latency on a MacBook Pro (Apple M1 Pro, 8-core CPU, 16GB RAM) using three MLPs and three CNNs of increasing complexity, summarized in Table~\ref{tab:models}.
Since our evaluation focuses exclusively on inference latency rather than predictive performance, we use synthetic inputs that match the dimensionality and shape of the CIFAR-10 dataset. For each model, we compare plain PyTorch inference with SMPC-based inference under varying batch sizes.
To isolate pure computational overhead, we report results for single-party (1P) execution, as well as multi-party execution using local multiprocessing to emulate distributed SMPC parties.
All reported latency measurements are averaged over three independent runs.

\paragraph{Baseline Inference Cost.}
Plain PyTorch inference on CPU achieves millisecond-level latency across all tested models and batch sizes (Table~\ref{tab:latency_1p}).
Specifically, even for convolutional models, inference latency remains within single-digit to low double-digit milliseconds.
These measurements serve as the baseline for quantifying the relative overhead introduced by the SMPC protocol in subsequent experiments.

\begin{table}[h]
\centering
\caption{Inference latency (ms) on CPU for three custom MLPs and CNNs under varying batch sizes (BS). Mean $\pm$ std over 3 runs.}
\label{tab:latency_1p}
\resizebox{\linewidth}{!}{
\begin{tabular}{c|cc|cc|cc}
\toprule
\multirow{2}{*}{BS} 
& \multicolumn{2}{c|}{\textbf{SmallMLP}} 
& \multicolumn{2}{c|}{\textbf{MediumMLP}} 
& \multicolumn{2}{c}{\textbf{LargeMLP}} \\
& Plain & MPC-1P 
& Plain & MPC-1P
& Plain & MPC-1P \\
\midrule
1 
& 0.0 $\pm$ 0.0 & 18.2 $\pm$ 0.6
& 0.1 $\pm$ 0.0 & 35.7 $\pm$ 0.2
& 0.4 $\pm$ 0.1 & 74.8 $\pm$ 0.3 \\
2 
& 0.1 $\pm$ 0.0 & 20.4 $\pm$ 1.7
& 0.6 $\pm$ 0.1 & 41.7 $\pm$ 2.2
& 1.2 $\pm$ 0.0 & 83.1 $\pm$ 1.5 \\
4 
& 0.1 $\pm$ 0.0 & 22.2 $\pm$ 0.3
& 0.6 $\pm$ 0.0 & 44.9 $\pm$ 0.4
& 1.2 $\pm$ 0.1 & 94.8 $\pm$ 1.9 \\
8 
& 0.1 $\pm$ 0.0 & 28.0 $\pm$ 1.0
& 0.5 $\pm$ 0.1 & 57.0 $\pm$ 0.8
& 1.3 $\pm$ 0.2 & 118.8 $\pm$ 0.3 \\
16 
& 0.2 $\pm$ 0.0 & 38.8 $\pm$ 0.3
& 0.6 $\pm$ 0.0 & 81.6 $\pm$ 1.6
& 1.1 $\pm$ 0.0 & 172.2 $\pm$ 1.6 \\
32 
& 0.2 $\pm$ 0.0 & 63.4 $\pm$ 0.6
& 0.6 $\pm$ 0.0 & 127.7 $\pm$ 1.0
& 0.9 $\pm$ 0.1 & 275.5 $\pm$ 1.9 \\
64 
& 0.4 $\pm$ 0.1 & 106.3 $\pm$ 0.8
& 0.6 $\pm$ 0.1 & 221.1 $\pm$ 0.9
& 1.8 $\pm$ 0.8 & 512.5 $\pm$ 23.5 \\

\toprule
\multirow{2}{*}{BS} 
& \multicolumn{2}{c|}{\textbf{SmallCNN}} 
& \multicolumn{2}{c|}{\textbf{MediumCNN}} 
& \multicolumn{2}{c}{\textbf{LargeCNN}} \\
& Plain & MPC-1P 
& Plain & MPC-1P
& Plain & MPC-1P \\
\midrule
1 
& 0.5 $\pm$ 0.2 & 94.2 $\pm$ 2.7
& 2.1 $\pm$ 0.3 & 208.8 $\pm$ 7.1
& 3.4 $\pm$ 0.3 & 410.7 $\pm$ 34.3 \\
2 
& 1.9 $\pm$ 0.7 & 124.7 $\pm$ 3.6
& 2.7 $\pm$ 0.2 & 257.2 $\pm$ 2.5
& 6.1 $\pm$ 0.3 & 520.5 $\pm$ 16.5 \\
4 
& 1.6 $\pm$ 0.1 & 169.8 $\pm$ 1.3
& 4.2 $\pm$ 0.2 & 350.4 $\pm$ 6.1
& 7.7 $\pm$ 0.3 & 667.9 $\pm$ 11.8 \\
8 
& 1.9 $\pm$ 0.2 & 289.1 $\pm$ 13.3
& 5.1 $\pm$ 0.3 & 533.4 $\pm$ 3.3
& 10.9 $\pm$ 0.5 & 1047.1 $\pm$ 26.9 \\
16 
& 5.5 $\pm$ 0.6 & 461.1 $\pm$ 3.8
& 12.0 $\pm$ 0.6 & 928.1 $\pm$ 12.5
& 32.2 $\pm$ 0.7 & 1771.4 $\pm$ 19.5 \\
32 
& 9.3 $\pm$ 0.3 & 913.8 $\pm$ 31.6
& 17.8 $\pm$ 0.6 & 1614.6 $\pm$ 34.5
& 47.9 $\pm$ 1.3 & 3345.9 $\pm$ 69.0 \\
64 
& 15.2 $\pm$ 0.2 & 1642.2 $\pm$ 19.3
& 29.2 $\pm$ 1.5 & 3114.4 $\pm$ 73.3
& 72.9 $\pm$ 1.7 & 6905.5 $\pm$ 622.0 \\
\bottomrule
\end{tabular}
}
\end{table}

\paragraph{Computation Overhead under 1P SMPC Execution.}
CrypTen-based SMPC inference incurs substantially higher latency even under single-party execution.
Although 1P execution involves no inter-party communication, all computations are still carried out using MPC-specific primitives, including secret-shared representations, fixed-point arithmetic, and secure implementations of non-linear functions. As a result, inference latency increases by one to two orders of magnitude compared to plain execution.
For example, CNN-based models exceed 90~ms latency at batch size~1 and reach several hundred milliseconds as batch size increases (Table~\ref{tab:latency_1p}).
Overall, 1P SMPC introduces approximately 50×–200× latency overhead compared to plain inference, depending on model complexity and batch size.
These results indicate that a substantial portion of SMPC overhead already arises from protocol-induced computation costs, even in the absence of network communication.

\paragraph{Multi-Party Execution.}
When extending to cryptographically secure multi-party configurations, inference latency increases further.
As shown in Table~\ref{tab:latency_multiparty}, moving from two to five parties shifts the latency from the sub-second regime to multiple seconds.
It is important to note that this latency growth is not an artifact of system-level resource contention or inter-process communication delays, as sufficient computational resources were allocated to execute parties in parallel.
Instead, the increase is intrinsic to the SMPC protocol itself: as the number of parties grows, the volume of arithmetic operations required to manage and aggregate secret shares inherently scales, leading to higher computational costs.
\paragraph{Summary.}
Overall, these results establish a lower bound on inference latency for
SMPC-based execution that arises purely from computation.
Even without network communication, SMPC inference operates in a
latency regime that is substantially higher than plain inference.
The effect of network communication and geographically distributed
deployment is examined separately in the next subsection. This decomposition allows us to later attribute additional latency observed in distributed settings specifically to network communication rather than SMPC computation itself.

\begin{table}[h]
\centering
\caption{Inference latency (ms) on CPU for SmallMLP and SmallCNN using CrypTen-$N$P. Mean $\pm$ std over 3 runs.}
\label{tab:latency_multiparty}
\resizebox{\linewidth}{!}{%
\begin{tabular}{c c c c c c}
\toprule
\textbf{MLP} & bs=1 & bs=2 & bs=4 & bs=8 & bs=16 \\
\midrule
Party=2 & 71.9 $\pm$ 2.3 & 74.4 $\pm$ 0.8 & 85.3 $\pm$ 1.5 & 106.2 $\pm$ 1.9 & 130.9 $\pm$ 0.9 \\
Party=3 & 108.3 $\pm$ 1.2 & 115.5 $\pm$ 9.0 & 123.0 $\pm$ 1.0 & 142.0 $\pm$ 1.1 & 182.7 $\pm$ 0.7 \\
Party=4 & 158.4 $\pm$ 0.4 & 164.8 $\pm$ 3.3 & 170.7 $\pm$ 1.3 & 189.3 $\pm$ 2.5 & 225.6 $\pm$ 1.1 \\
Party=5 & 267.5 $\pm$ 1.5 & 271.9 $\pm$ 0.4 & 274.1 $\pm$ 5.0 & 293.2 $\pm$ 5.3 & 328.5 $\pm$ 2.7 \\
\midrule
\textbf{CNN} & bs=1 & bs=2 & bs=4 & bs=8 & bs=16 \\
\midrule
Party=2 & 466.4 $\pm$ 9.2 & 724.5 $\pm$ 9.9 & 1207.4 $\pm$ 19.0 & 2149.1 $\pm$ 62.3 & 4606.5 $\pm$ 70.8 \\
Party=3 & 1231.8 $\pm$ 22.5 & 1798.5 $\pm$ 14.4 & 2843.9 $\pm$ 13.1 & 5804.1 $\pm$ 61.2 & 11724.8 $\pm$ 187.3 \\
Party=4 & 1810.9 $\pm$ 22.8 & 2918.8 $\pm$ 110.1 & 5185.4 $\pm$ 227.1 & 9893.2 $\pm$ 102.8 & 20905.0 $\pm$ 66.6 \\
Party=5 & 3271.6 $\pm$ 16.5 & 4808.3 $\pm$ 91.4 & 8302.3 $\pm$ 81.7 & 16313.2 $\pm$ 324.1 & 33026.0 $\pm$ 376.9 \\
\bottomrule
\end{tabular}
}
\end{table}

\begin{figure*}
    \centering
    \includegraphics[width=\linewidth]{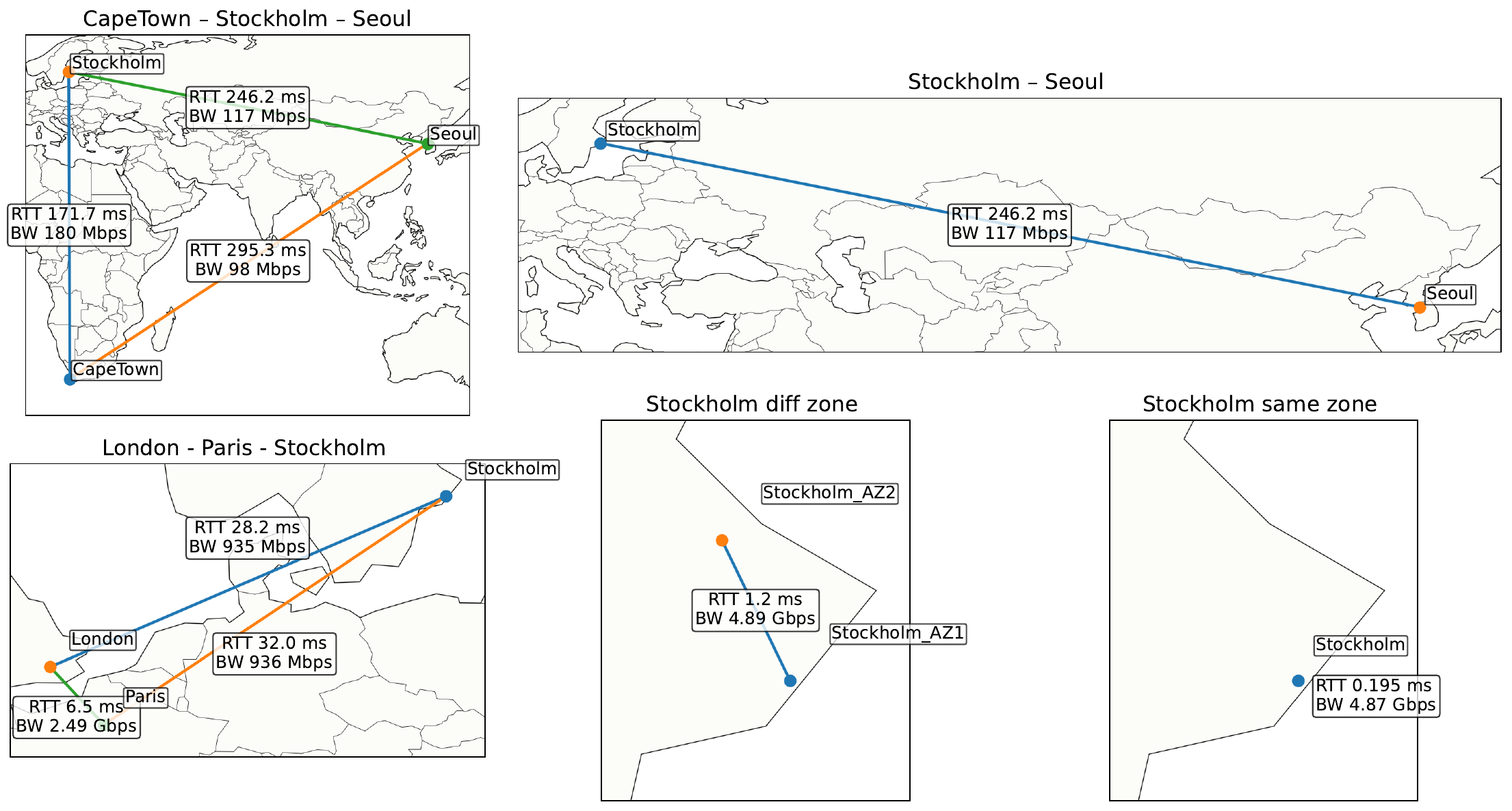}
    \caption{Representative network deployment scenarios used in the evaluation. Node locations are shown schematically for illustration and do not reflect exact locations. Each link is annotated by RTT and available bandwidth (BW).}
    \label{fig:latency}
\end{figure*}

\begin{table*}
\centering
\caption{FI latency under different network settings and corresponding network characteristics.
Latency is reported in seconds as mean $\pm$ std over three runs.
$K$ denotes the number of models participating in ensemble inference, while the number of MPC parties is fixed to three.
Batch size is set to 1.}

\label{tab:fi_latency_network_combined}
\resizebox{\textwidth}{!}{
\begin{tabular}{llcc|cc|cc}
\toprule
\textbf{Case} & \textbf{Network Setting} 
& \textbf{RTT (ms)} & \textbf{Bandwidth}
& \multicolumn{2}{c|}{\textbf{LeNet Latency (s)}} 
& \multicolumn{2}{c}{\textbf{ResNet-18 Latency (s)}} \\
\cmidrule(lr){5-6} \cmidrule(lr){7-8}
& & & 
& $K=1$ & $K=3$
& $K=1$ & $K=3$ \\
\midrule
1 & Intra-zone (Stockholm)
& $\sim$0.20 & $\sim$4.9 Gbps
& $0.31 \pm 0.01$ & $0.96 \pm 0.02$
& $6.16 \pm 0.03$ & $18.54 \pm 0.02$ \\

2 & Inter-zone (Stockholm regions)
& $\sim$1.16 & $\sim$4.9 Gbps
& $0.59 \pm 0.00$ & $1.91 \pm 0.12$
& $7.11 \pm 0.07$ & $21.43 \pm 0.12$ \\

3 & Multi-zone EU (London--Paris--Stockholm)
& 6--32 & 0.9--2.7 Gbps
& $9.13 \pm 0.07$ & $27.85 \pm 0.25$
& $43.88 \pm 0.65$ & $131.05 \pm 0.75$ \\

4 & Inter-continent (Seoul--Stockholm)
& $\sim$250 & $\sim$120 Mbps
& $72.67 \pm 0.44$ & $218.13 \pm 0.75$
& $309.91 \pm 1.54$ & $933.27 \pm 1.86$ \\

5 & Global (S. Africa--Sweden--S. Korea)
& 170--295 & 95--180 Mbps
& $83.62 \pm 1.32$ & $263.09 \pm 2.12$
& $365.97 \pm 4.80$ & $1085.97 \pm 5.12$ \\
\bottomrule
\end{tabular}
}
\end{table*}

\subsection{Impact of Geographic Network Latency}\label{sec:network_overhead}

\paragraph{Motivation and Scope.}
This experiment examines how geographic network conditions affect the end-to-end latency of MPC-based FI. Our goal is not to isolate the individual impact of round-trip time (RTT) or bandwidth, but to establish a practical latency benchmark using representative deployment scenarios. By providing empirical references for familiar model architectures (e.g., LeNet and ResNet), we aim to offer a decision-making framework for organizations assessing the feasibility of collaborative inference relative to their geographic distribution.

\paragraph{Experimental Setup.}
All experiments are conducted on Amazon Web Services (AWS) using \texttt{t3.micro} instances\footnote{To ensure consistent computational performance, we explicitly enabled the `Unlimited' mode for T3 instances. This configuration prevents CPU throttling by allowing instances to sustain high CPU performance beyond the baseline, ensuring that the reported latencies are not confounded by credit exhaustion.} (2 vCPUs, 1~GB memory). Parties are deployed across different regions and subnets to emulate co-located, regional, and cross-continent deployments, relying solely on AWS-provided networking\footnote{CrypTen relies on the Gloo backend for inter-party communication, which assumes connectivity over private IP addresses. Since parties are deployed across different AWS regions and subnets, we additionally configured VPC peering to enable private IP routing between instances, ensuring stable and direct inter-process communication without traversing public endpoints.}. We fix the batch size to 1 and the world size to 3, varying only the geographic placement of the participating parties. For each deployment, we report both inference latency and the corresponding network characteristics, including RTT and available bandwidth (Figure~\ref{fig:latency} and Table~\ref{tab:fi_latency_network_combined}).
\paragraph{Latency under Local and Regional Deployments.}
When all parties are co-located within the same availability zone in Stockholm, inference latency remains below one second for LeNet and within tens of seconds for ResNet-18. Moving to different availability zones within the same region introduces a modest increase in latency while preserving similar bandwidth. These results indicate that for regionally proximate participants, MPC-based inference is viable for near-real-time services, particularly for lightweight models.

\paragraph{Latency under Geographically Distributed Deployments.}
As parties are distributed across multiple European regions and continents, inference latency increases substantially. Multi-zone European deployments push latency into the order of tens of seconds, while inter-continent and global configurations result in latencies spanning several minutes. This trend holds for both LeNet and ResNet-18; while deeper models incur higher absolute latency, they exhibit similar scaling behavior relative to geographic distance.

\paragraph{Interpretation and Benchmark Value.}
These results reveal that under wide-area deployments, network-induced communication overhead dominates end-to-end latency, often overwhelming local computation costs. Our data shows that reducing computation overhead alone is insufficient to recover low-latency performance in such settings. Beyond confirming the intuitive impact of distance, this experiment establishes a quantitative baseline for the current state of SMPC-based FI. For practitioners, these findings serve as a prerequisite for feasibility studies, identifying the specific network regimes where real-time collaboration remains practical.

\paragraph{Summary and Future Guidance.}
In summary, our benchmarks demonstrate that SMPC-based collaborative inference is constrained by a dual bottleneck of computational overhead and geographic network latency. By providing empirical latency references for standard models across representative cloud deployments, this study provides a concrete baseline for future research aiming to develop more efficient MPC primitives or communication-optimized protocols. Ultimately, our findings highlight that geographic network conditions, rather than model complexity alone, impose the fundamental constraint on the practical scalability of SMPC-based collaborative systems.

\subsection{Ensemble Accuracy under Non-IID Data}
\label{subsec:ensemble_accuracy}

\paragraph{Motivation and Scope.}
This experiment evaluates the predictive performance of ensemble-based inference, which serves as the primary mechanism for collaboration in FedSEI.
Unlike conventional ensembles in centralized learning, where multiple models are trained on the same dataset to improve generalization, our setting assumes that each client trains its model independently on a distinct and potentially heterogeneous local dataset.
We investigate how different ensemble aggregation strategies behave under non-IID (non–independent and identically distributed) data distributions and client scaling, examining when and why collaborative inference yields performance gains over individual client models.

\begin{figure}
    \centering
    \includegraphics[width=\linewidth]{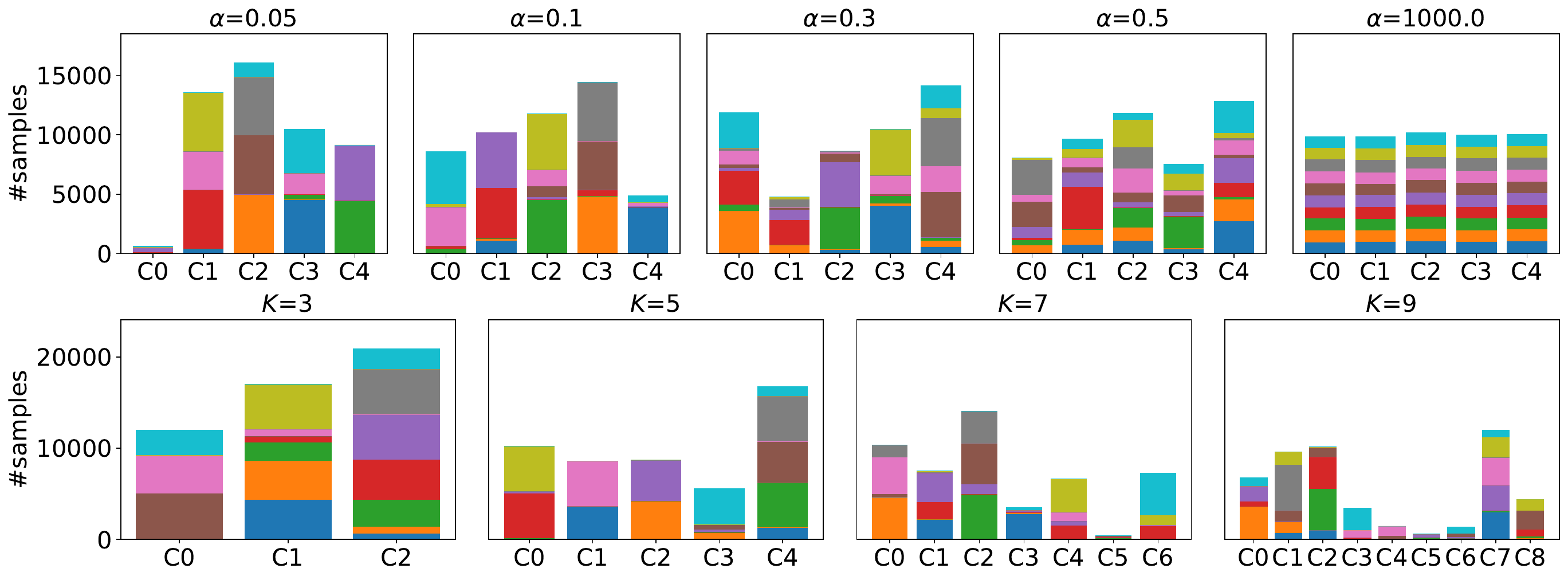}
    \caption{
Label distribution across clients under Dirichlet-based non-IID partitioning on CIFAR-10.
Each subplot shows a stacked bar chart where the x-axis denotes client indices,
the y-axis indicates the number of samples, and colors represent class labels.
(\textbf{Top}) Effect of varying the Dirichlet concentration parameter $\alpha$ with $K=5$.
Smaller $\alpha$ results in more heterogeneous and imbalanced client data.
(\textbf{Bottom}) Effect of varying the number of clients $K$ with $\alpha=0.1$,
leading to smaller per-client datasets while preserving non-IID characteristics.
}

    \label{fig:non-iid-distribution}
\end{figure}

\begin{table*}[t]
\centering
\caption{Ensemble performance under varying $K$ and $\alpha$.
Single Avg and Single Best denote the average and best accuracy of individual client models, respectively.
Hard and Soft correspond to hard voting and uniform probability averaging.
Results are reported as mean accuracy (standard deviation) over three random seeds, with the best and second-best ensemble results per row highlighted in \textbf{bold} and \underline{underline}.}

\label{tab:ensemble_alpha_K_full}
\resizebox{\linewidth}{!}{
\begin{tabular}{llcccccccc}
\toprule
Dataset / Model & $K$ & $\alpha$ & Single Avg & Single Best & Hard & Soft & Entropy & Spectral & TTA \\
\midrule
\multirow{20}{*}{\shortstack{CIFAR-10 \\ LeNet}} & \multirow{5}{*}{3} & 0.05 & 30.55 (1.57) & \underline{42.69 (2.92)} & 30.94 (8.75) & 41.18 (2.91) & 41.72 (2.75) & \textbf{43.01 (2.76)} & 42.57 (1.18) \\
  &   & 0.1 & 33.86 (3.55) & 43.52 (3.38) & 36.24 (5.21) & \underline{50.65 (1.24)} & \textbf{52.78 (4.24)} & 46.91 (3.21) & 45.71 (3.52) \\
  &   & 0.3 & 42.49 (0.23) & 48.17 (0.81) & 49.17 (1.08) & \underline{61.91 (0.41)} & \textbf{62.01 (1.03)} & 57.33 (1.80) & 56.40 (1.01) \\
  &   & 0.5 & 48.29 (2.66) & 53.44 (3.77) & 55.55 (3.16) & \underline{62.86 (0.68)} & \textbf{64.28 (0.13)} & 62.72 (0.64) & 60.19 (0.22) \\
  &   & 1000.0 & 60.83 (0.87) & 62.46 (1.77) & 63.69 (1.00) & \underline{65.63 (0.69)} & 65.53 (0.88) & \textbf{65.66 (0.68)} & 63.80 (0.54) \\
  \cmidrule{3-10}
  & \multirow{5}{*}{5} & 0.05 & 23.72 (1.68) & \underline{40.50 (3.20)} & 29.66 (10.94) & 28.60 (13.39) & 26.27 (12.93) & \textbf{45.27 (5.52)} & 37.28 (6.82) \\
  &   & 0.1 & 28.52 (2.18) & 41.74 (2.90) & 40.59 (2.20) & 42.54 (2.58) & 42.68 (5.16) & \underline{43.04 (3.38)} & \textbf{43.50 (3.03)} \\
  &   & 0.3 & 36.48 (2.06) & 41.48 (2.81) & 48.88 (1.91) & \underline{57.33 (0.95)} & \textbf{59.12 (1.66)} & 51.47 (4.11) & 52.55 (1.40) \\
  &   & 0.5 & 41.39 (2.03) & 50.28 (1.71) & 54.04 (1.28) & \underline{60.31 (1.15)} & \textbf{61.76 (0.65)} & 59.45 (0.49) & 58.26 (1.25) \\
  &   & 1000.0 & 55.99 (0.98) & 57.95 (0.76) & 60.66 (1.06) & \underline{61.81 (0.67)} & 61.74 (0.45) & \textbf{61.88 (0.57)} & 60.63 (0.42) \\
  \cmidrule{3-10}
  & \multirow{5}{*}{7} & 0.05 & 20.19 (1.31) & 26.91 (1.23) & 31.84 (3.87) & 32.19 (6.25) & 30.14 (7.88) & \textbf{34.01 (3.97)} & \underline{32.47 (1.97)} \\
  &   & 0.1 & 22.77 (2.02) & 33.62 (5.47) & 36.89 (1.56) & \underline{42.25 (1.31)} & \textbf{45.26 (4.90)} & 38.08 (3.99) & 40.11 (2.07) \\
  &   & 0.3 & 32.19 (1.11) & 42.75 (3.04) & 46.45 (1.40) & \underline{53.10 (2.43)} & \textbf{54.94 (1.75)} & 48.47 (2.87) & 49.89 (1.66) \\
  &   & 0.5 & 37.01 (1.94) & 47.86 (3.74) & 52.51 (1.60) & \underline{57.68 (1.04)} & \textbf{58.53 (1.26)} & 56.91 (1.87) & 55.30 (0.39) \\
  &   & 1000.0 & 53.45 (1.01) & 56.04 (0.34) & 58.75 (1.23) & \underline{59.58 (1.20)} & 59.57 (1.27) & \textbf{59.62 (1.19)} & 58.65 (1.16) \\

\midrule
\multirow{20}{*}{\shortstack{CIFAR-100 \\ ResNet-18}} & \multirow{5}{*}{3} & 0.05 & 26.50 (1.03) & 31.72 (0.98) & 28.75 (1.46) & \underline{49.48 (1.38)} & \textbf{50.99 (1.41)} & 45.77 (0.98) & 34.15 (0.67) \\
  &   & 0.1 & 28.89 (0.55) & 32.74 (1.04) & 31.22 (1.47) & \underline{52.53 (0.16)} & \textbf{53.86 (0.32)} & 46.56 (3.31) & 36.70 (0.41) \\
  &   & 0.3 & 35.00 (1.01) & 36.27 (1.91) & 40.94 (1.18) & \underline{54.77 (1.66)} & \textbf{55.31 (1.83)} & 54.09 (0.97) & 40.86 (1.15) \\
  &   & 0.5 & 38.04 (1.49) & 41.43 (3.49) & 45.11 (2.00) & \textbf{55.87 (2.05)} & \underline{55.78 (1.94)} & 55.46 (2.07) & 43.33 (0.13) \\
  &   & 1000.0 & 47.73 (1.55) & 50.02 (0.95) & 52.67 (1.64) & \textbf{56.74 (1.37)} & 55.97 (1.46) & \underline{56.72 (1.40)} & 49.14 (1.27) \\
  \cmidrule{3-10}
  & \multirow{5}{*}{5} & 0.05 & 16.92 (0.18) & 20.22 (1.05) & 21.58 (0.65) & \underline{40.42 (1.06)} & \textbf{42.41 (1.25)} & 35.37 (2.55) & 26.29 (1.63) \\
  &   & 0.1 & 19.55 (0.37) & 23.99 (1.65) & 28.09 (0.60) & \underline{43.02 (0.67)} & \textbf{45.14 (0.93)} & 40.75 (0.79) & 30.11 (1.00) \\
  &   & 0.3 & 26.26 (1.50) & 29.98 (1.01) & 39.67 (2.31) & \underline{49.03 (2.13)} & \textbf{49.23 (2.21)} & 48.84 (2.17) & 37.37 (1.06) \\
  &   & 0.5 & 28.90 (0.54) & 32.75 (1.05) & 42.63 (0.88) & \textbf{50.24 (0.29)} & \underline{49.82 (0.33)} & 49.74 (0.40) & 39.30 (0.70) \\
  &   & 1000.0 & 38.69 (0.48) & 41.45 (0.55) & 48.08 (0.65) & \underline{51.17 (0.52)} & 49.95 (0.43) & \textbf{51.23 (0.52)} & 45.00 (0.85) \\
  \cmidrule{3-10}
  & \multirow{5}{*}{7} & 0.05 & 12.96 (0.47) & 17.52 (0.31) & 20.00 (1.95) & \underline{34.64 (1.54)} & \textbf{36.86 (0.89)} & 31.18 (2.26) & 23.82 (1.36) \\
  &   & 0.1 & 15.46 (0.19) & 19.23 (0.91) & 27.61 (1.26) & \underline{38.99 (1.11)} & \textbf{40.78 (0.87)} & 36.84 (0.87) & 27.89 (0.19) \\
  &   & 0.3 & 20.28 (0.20) & 24.13 (0.30) & 35.90 (0.63) & \textbf{43.23 (1.00)} & \underline{43.12 (0.75)} & 43.12 (0.64) & 34.74 (0.80) \\
  &   & 0.5 & 23.36 (0.93) & 29.35 (3.21) & 39.14 (1.62) & \textbf{45.90 (1.40)} & 45.22 (1.28) & \underline{45.50 (1.59)} & 37.86 (0.36) \\
  &   & 1000.0 & 32.32 (0.76) & 35.96 (0.50) & 43.23 (0.94) & \underline{46.04 (0.78)} & 44.69 (0.82) & \textbf{46.06 (0.78)} & 41.93 (0.85) \\
  
\bottomrule
\end{tabular}
}
\end{table*}

\paragraph{Datasets.}
We evaluate FedSEI on six benchmark datasets spanning standard computer vision and medical imaging tasks.
These include CIFAR-10 and CIFAR-100~\cite{krizhevsky2009learning} (RGB images with 10 and 100 classes), Fashion-MNIST~\cite{xiao2017fashion} (grayscale clothing images with 10 classes), and EMNIST~\cite{cohen2017emnist} (grayscale handwritten characters with 47 classes).
We additionally include two medical datasets from MedMNIST~\cite{medmnistv1}: PathMNIST (colon pathology images with 9 classes) and OrganAMNIST (abdominal CT images with 11 classes).
All datasets follow their standard train/validation/test splits.

\paragraph{Non-IID Data and Client Scaling.}
We model data heterogeneity using a class-wise Dirichlet distribution controlled by parameter $\alpha$~\cite{seo2025understanding}.
Larger values of $\alpha$ correspond to more balanced, IID-like data partitions, while smaller values induce stronger label skew.
Figure~\ref{fig:non-iid-distribution} illustrates representative non-IID data allocations under Dirichlet partitioning. This setup reflects realistic cross-silo collaboration scenarios, where participants are expected to hold complementary data distributions; otherwise, the incentive for collaborative inference would be limited.

\paragraph{Experimental Setup.}
Each client trains its local model independently using only its allocated data.
We apply a $9{:}1$ train--validation split within each local dataset and train models for up to 50 epochs with early stopping.
Training is terminated if the validation accuracy does not improve for five consecutive epochs, and the checkpoint achieving the highest validation accuracy is selected for inference.
Unless otherwise stated, all models are trained with a batch size of 128 using the Adam optimizer with a learning rate of 0.01 and a weight decay of $5\times10^{-4}$.
All hyperparameters are kept fixed across experiments to isolate the effects of data heterogeneity and client scaling.

\paragraph{Ensemble Strategies.}
We evaluate hard voting and soft voting as uniform ensemble baselines.
Beyond uniform averaging, we examine whether simple, label-free adjustments to ensemble weights can provide performance headroom in inference-time collaboration, where ground-truth labels are unavailable.
The objective of this study is not to propose new ensemble algorithms or to analyze the optimality of specific heuristics, but to empirically probe the feasibility of query-dependent weighting within the FI setting.

To this end, we consider several lightweight label-free weighting schemes.
We employ entropy-based weighting to reflect prediction uncertainty.
We additionally include two exploratory heuristics adapted for non-IID environments: (i) a spectral weighting scheme inspired by SUMMA~\cite{ahsen2019unsupervised}, tailored to estimate relative model reliability from confidence covariance, and (ii) a test-time augmentation (TTA)-based scheme~\cite{ayhan2018test}.
For the TTA-based scheme, we further test a simple hypothesis for label-skewed settings by assigning higher weights to models that exhibit larger prediction variation under input perturbations.
Unlike static averaging, these methods allow ensemble weights to vary across queries, serving as probes for the potential benefit of dynamic collaboration.
Implementation details and experimental configurations for the label-free weighting schemes considered in this study, including TTA settings and entropy computation, are provided in Appendix~\ref{app:weighting_details}.
\footnote{All ensemble methods are SMPC-compatible, with the implementation available in our repository.}

\begin{figure}
    \centering
    \includegraphics[width=\linewidth]{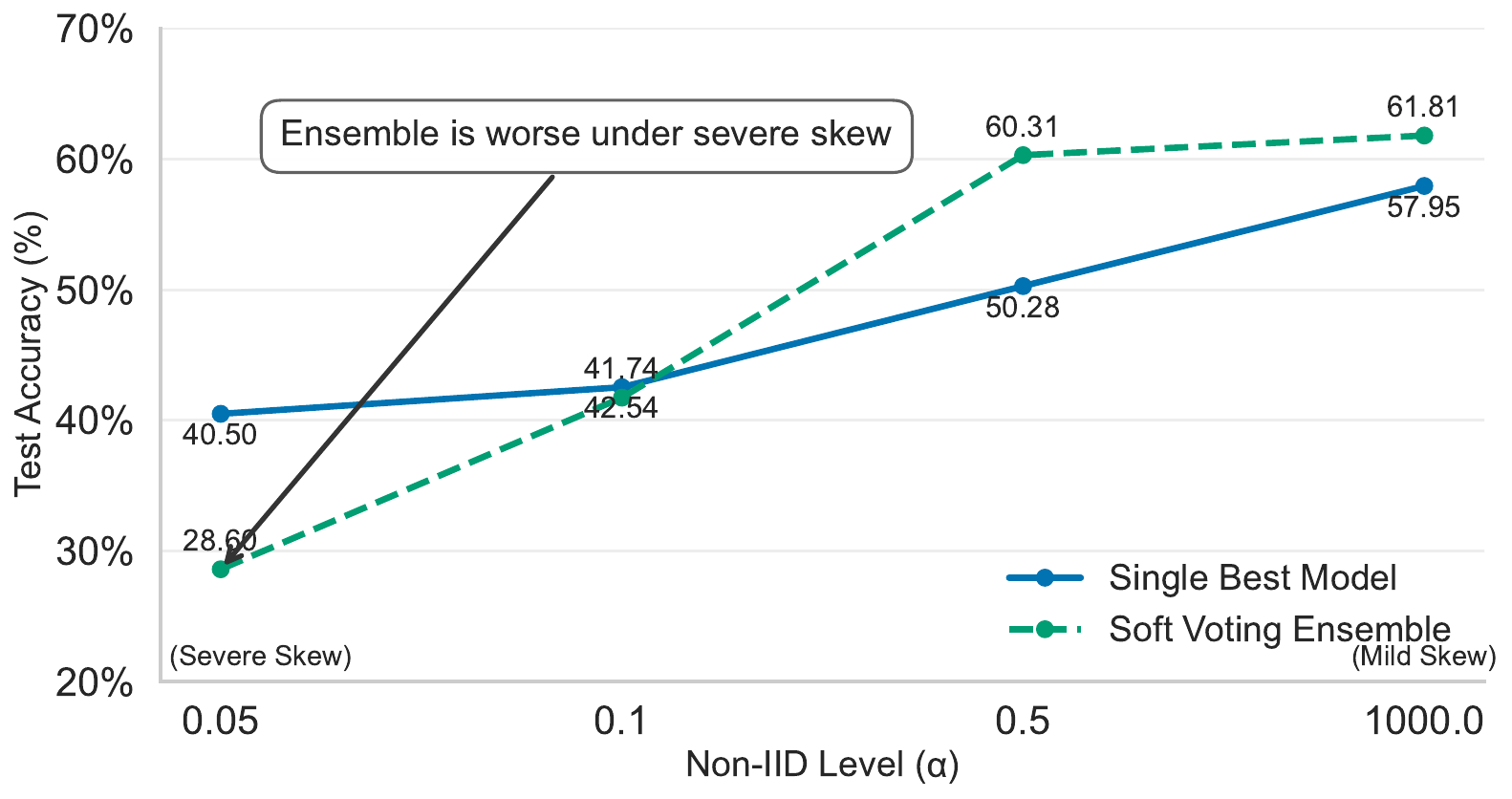}
    \caption{Test accuracy of a single best model and a soft-voting ensemble under different non-IID levels on CIFAR-10 (K=5).}
    \label{fig:ensemble-underperformance-example}
\end{figure}

\begin{table*}[t]
\centering
\caption{Ensemble performance with ResNet-18 under $K=3$ and varying $\alpha$ for different datasets. Results are reported as mean accuracy (standard deviation) over three random seeds, with the best and second-best ensemble results per row highlighted in \textbf{bold} and \underline{underline}.}
\label{tab:ensemble_alpha_K_ResNet-18}
\resizebox{\linewidth}{!}{
\begin{tabular}{llcccccccc}
\toprule
Dataset / Model & $K$ & $\alpha$ & Single Avg & Single Best & Hard & Soft & Entropy & Spectral & TTA \\
\midrule
\multirow{5}{*}{\shortstack{EMNIST \\ ResNet-18}} & \multirow{5}{*}{3} & 0.05 & 39.29 (0.14) & 45.08 (2.89) & 47.95 (3.60) & \underline{71.54 (1.05)} & \textbf{74.86 (1.01)} & 47.98 (6.80) & 56.50 (1.44) \\
  &   & 0.1 & 46.91 (2.85) & 52.33 (2.56) & 57.43 (3.14) & \underline{72.59 (0.95)} & \textbf{76.01 (0.84)} & 50.84 (5.89) & 64.78 (1.00) \\
  &   & 0.3 & 61.08 (3.03) & 66.53 (3.46) & 75.45 (3.61) & \underline{80.86 (1.69)} & \textbf{82.23 (0.91)} & 68.49 (6.52) & 77.64 (1.69) \\
  &   & 0.5 & 70.08 (1.80) & 74.96 (2.88) & 81.31 (0.93) & \underline{83.24 (0.86)} & \textbf{83.76 (0.51)} & 83.05 (0.72) & 81.22 (1.27) \\
  &   & 1000.0 & 83.56 (0.47) & 83.94 (0.35) & 85.44 (0.36) & \textbf{86.05 (0.21)} & 86.02 (0.21) & \underline{86.05 (0.24)} & 85.19 (0.40) \\

\midrule
\multirow{5}{*}{\shortstack{Fashion-MNIST \\ ResNet-18}} & \multirow{5}{*}{3} & 0.05 & 43.28 (4.67) & 56.55 (10.88) & 44.66 (9.96) & 57.27 (6.70) & 58.59 (9.35) & \textbf{66.29 (8.98)} & \underline{63.29 (8.22)} \\
  &   & 0.1 & 48.70 (1.45) & 65.94 (10.26) & 54.17 (6.00) & \underline{72.10 (9.83)} & \textbf{76.58 (7.04)} & 50.44 (6.41) & 69.36 (10.44) \\
  &   & 0.3 & 72.31 (3.84) & 78.88 (7.40) & 83.78 (4.63) & \underline{86.52 (2.86)} & \textbf{87.34 (2.40)} & 74.82 (14.05) & 85.38 (2.34) \\
  &   & 0.5 & 76.97 (1.40) & 81.78 (1.08) & 87.32 (3.87) & \textbf{89.21 (1.89)} & \underline{88.76 (2.53)} & 84.52 (4.84) & 87.77 (1.51) \\
  &   & 1000.0 & 89.21 (0.84) & 90.11 (0.16) & 90.86 (0.51) & \textbf{91.22 (0.56)} & 91.16 (0.61) & \underline{91.20 (0.61)} & 90.51 (0.55) \\

\midrule
\multirow{5}{*}{\shortstack{PathMNIST \\ ResNet-18}} & \multirow{5}{*}{3} & 0.05 & 42.16 (2.49) & 55.52 (6.80) & 50.03 (3.58) & \underline{61.85 (9.65)} & \textbf{65.47 (12.71)} & 41.11 (5.93) & 61.01 (8.64) \\
  &   & 0.1 & 49.32 (3.48) & 57.73 (1.15) & 60.01 (13.05) & \underline{64.48 (7.81)} & \textbf{69.19 (4.89)} & 58.26 (4.02) & 63.55 (6.12) \\
  &   & 0.3 & 57.92 (4.55) & 72.72 (5.15) & 70.45 (7.18) & \underline{74.53 (5.74)} & \textbf{75.91 (4.14)} & 65.40 (11.61) & 72.24 (4.26) \\
  &   & 0.5 & 69.61 (1.11) & 78.80 (0.65) & 80.90 (2.83) & \underline{84.36 (2.16)} & \textbf{85.94 (2.28)} & 74.50 (3.91) & 82.49 (0.83) \\
  &   & 1000.0 & 80.31 (3.87) & \underline{84.77 (2.44)} & 82.73 (6.31) & 84.46 (5.79) & \textbf{84.96 (5.41)} & 83.40 (6.00) & 83.94 (5.61) \\

\midrule
\multirow{5}{*}{\shortstack{OrganAMNIST \\ ResNet-18}} & \multirow{5}{*}{3} & 0.05 & 45.78 (3.46) & 54.10 (1.80) & 55.63 (11.38) & \underline{70.80 (1.35)} & \textbf{75.14 (1.26)} & 54.02 (11.31) & 64.85 (4.42) \\
  &   & 0.1 & 44.86 (2.46) & 51.01 (3.65) & 46.49 (10.32) & \underline{62.75 (6.41)} & \textbf{69.57 (6.65)} & 52.74 (12.23) & 54.21 (1.83) \\
  &   & 0.3 & 69.66 (2.90) & 77.39 (4.05) & 81.13 (1.83) & \underline{86.52 (0.51)} & \textbf{88.14 (0.33)} & 77.85 (3.26) & 83.33 (1.48) \\
  &   & 0.5 & 78.77 (1.40) & 84.58 (2.29) & 88.08 (1.22) & \underline{89.78 (0.35)} & \textbf{90.19 (0.46)} & 89.38 (0.77) & 87.02 (1.90) \\
  &   & 1000.0 & 89.37 (0.28) & 89.78 (0.23) & 91.68 (0.19) & 91.96 (0.23) & \underline{91.98 (0.27)} & \textbf{92.00 (0.18)} & 90.93 (0.33) \\

\bottomrule

\end{tabular}
}
\end{table*}

\paragraph{Results and Implications.}
Table~\ref{tab:ensemble_alpha_K_full} summarizes the performance of label-free ensemble strategies under varying degrees of data heterogeneity and client scaling.
Across many settings, \textit{Soft Voting} and \textit{Entropy-based weighting} serve as strong and reliable baselines, particularly on datasets with larger label spaces such as CIFAR-100, where the effects of Dirichlet-induced label skew are naturally mitigated~\cite{seo2026gc}.

However, under severe non-IID conditions, ensemble aggregation can become fragile.
On CIFAR-10 with $\alpha=0.05$, standard ensemble methods may underperform strong local models; for example, Soft Voting achieves only $28.60\%$ accuracy compared to $40.50\%$ for the \textit{Single Best} local model using LeNet with $K=5$.
This result indicates that naive aggregation does not guarantee performance improvements in highly skewed regimes (Figure~\ref{fig:ensemble-underperformance-example}).

Similar trends are observed beyond CIFAR-10 and CIFAR-100.
As shown in Table~\ref{tab:ensemble_alpha_K_ResNet-18}, experiments on additional datasets using ResNet-18 exhibit consistent patterns: while Soft Voting and Entropy-based weighting perform well overall, no single ensemble strategy emerges as a universal best.
In strongly non-IID settings, some of the considered non-uniform weighting schemes (e.g., entropy- or TTA-based weighting) occasionally achieve the highest accuracy, suggesting that query-dependent aggregation can be beneficial in specific regimes. 

Overall, these results indicate that ensemble effectiveness in FI is highly context-dependent.
While uniform aggregation provides a robust default, the varying success of different weighting strategies highlights the need for adaptive ensemble mechanisms that account for the degree of data heterogeneity and dataset characteristics, rather than relying on static aggregation rules.

\subsection{Label-Free Reward Allocation}
\label{subsec:reward_fairness}

\begin{figure*}
    \centering
    \includegraphics[width=\linewidth]{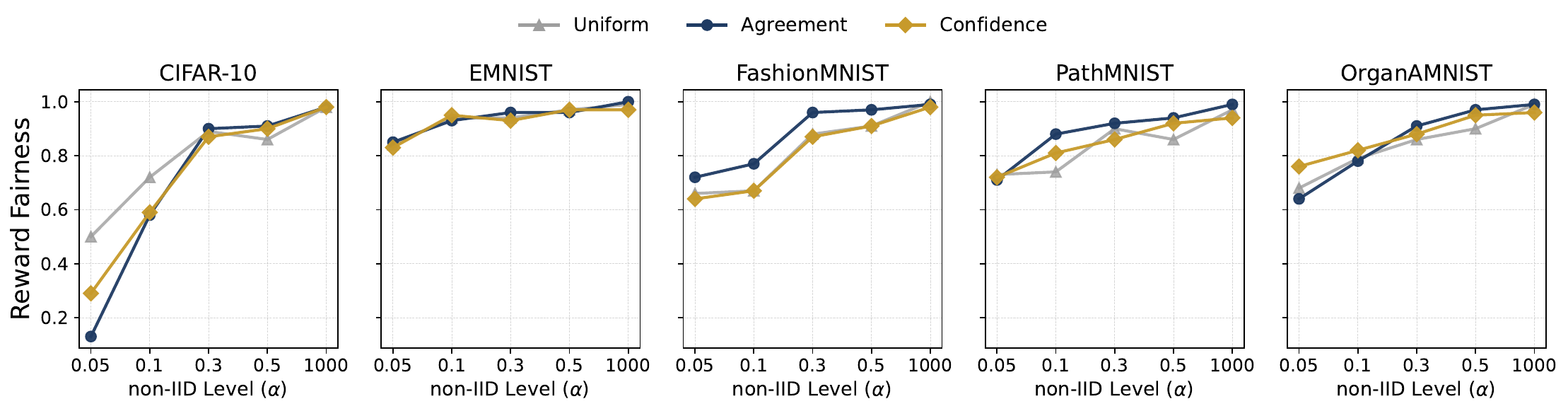}
    \caption{Reward fairness as a function of the non-IID level $\alpha$ across different datasets.
Higher values indicate fairer reward allocation.
Results are shown for five clients using LeNet models trained on heterogeneous local datasets.}

    \label{fig:reward-fairness-dataset}
\end{figure*}

\begin{figure*}
    \centering
    \includegraphics[width=\linewidth]{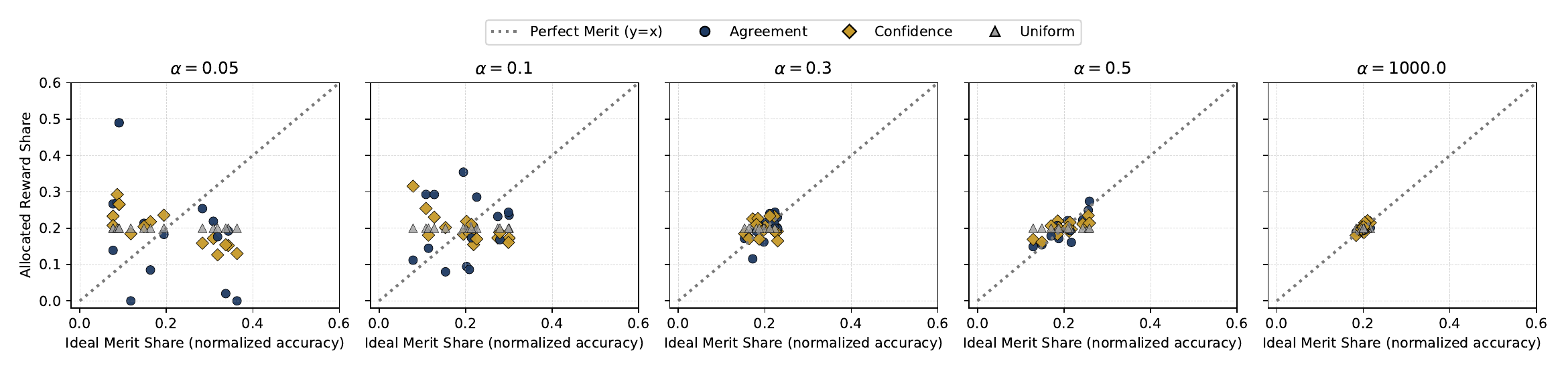}
    \caption{
Reward fairness analysis under different data heterogeneity levels.
The x-axis shows the \emph{ideal merit share} of each model, computed from normalized test accuracy and serving as an oracle reference.
The y-axis reports the \emph{allocated reward share} produced by label-free incentive schemes.
The diagonal line ($y=x$) represents perfect merit alignment, where rewards are proportional to true model quality.
Each point corresponds to one model across multiple random seeds.
Under strongly non-IID settings (small~$\alpha$), agreement- and confidence-based schemes exhibit substantial deviations from the ideal merit line and do not consistently improve upon uniform allocation.
As the data distribution becomes more IID-like (large~$\alpha$), all schemes converge toward merit-aligned behavior, reducing incentive ambiguity.
}

    \label{fig:reward-fairness-non-iid}
\end{figure*}

\paragraph{Motivation.}
Incentive mechanisms are essential for sustaining FI, as they directly affect participants’ willingness to contribute high-quality models.
A reward allocation scheme should ideally reflect each participant’s contribution, yet this goal is difficult to achieve in FI settings where neither ground-truth labels nor internal model information is available at inference time. In this section, we assume that each client acts as a model owner, i.e., the client submitting inference requests is also the provider of the corresponding smodel used for evaluation.

\paragraph{Ideal Merit-Based Allocation.}
As a reference point, we define an \emph{ideal merit share} based on individual model performance.
Let $a_k$ denote the test accuracy of client $k$, and let $K$ be the number of participating parties.
The ideal merit share is defined as
\begin{equation}
m_k = \frac{a_k}{\sum_{j=1}^{K} a_j},
\end{equation}
where $\sum_{k=1}^{K} m_k = 1$.
While this notion of fairness is not universally optimal, it provides a useful oracle for evaluating reward allocation schemes in a controlled setting. We emphasize that this merit definition is not claimed to be the correct objective for reward allocation, 
but rather serves as an oracle reference to probe how well label-free incentive schemes recover performance-proportional rewards in a controlled setting.

\paragraph{Label-Free Reward Allocation Schemes.}
In the absence of labels, reward allocation must rely solely on model outputs.
We consider three representative label-free schemes.

\emph{Uniform allocation} assigns equal reward to all clients:
\begin{equation}
r_k^{\text{uni}} = \frac{1}{K}.
\end{equation}

\emph{Confidence-based allocation} assigns higher reward to models producing more confident predictions.
Let $p_{k,n}$ denote the predicted probability vector of model $k$ for sample $n$, and let
\begin{equation}
H_{k,n} = - \sum_{c} p_{k,n}^{(c)} \log p_{k,n}^{(c)}
\end{equation}
be the prediction entropy.
The reward share is computed as
\begin{equation}
r_k^{\text{conf}} =
\frac{\sum_{n} \exp(-H_{k,n})}
{\sum_{j=1}^{K} \sum_{n} \exp(-H_{j,n})}.
\end{equation}

\emph{Agreement-based allocation} rewards models whose predictions agree with the ensemble output.
Let $y_n^{\text{ens}}$ denote the ensemble prediction for sample $n$.
The agreement score for client $k$ is defined as
\begin{equation}
A_k = \sum_{n} \mathbb{I}\bigl(y_{k,n} = y_n^{\text{ens}}\bigr),
\end{equation}
and the corresponding reward share is
\begin{equation}
r_k^{\text{agr}} = \frac{A_k}{\sum_{j=1}^{K} A_j}.
\end{equation}

\paragraph{Fairness Metric.}
To evaluate how closely a label-free reward allocation approximates the ideal merit-based allocation, we compare the normalized reward vector $\mathbf{r}$ with the ideal merit vector $\mathbf{m}$, where both vectors are normalized to sum to one.
We define \emph{Reward Fairness} as
\begin{equation}
\mathcal{F}(\mathbf{r}, \mathbf{m})
= 1 - \frac{1}{2} \left\lVert \mathbf{r} - \mathbf{m} \right\rVert_1,
\end{equation}
where higher values indicate closer alignment with merit-based allocation.
This metric allows us to assess whether label-free schemes improve upon uniform allocation. We adopt the $\ell_1$ distance due to its simplicity and direct interpretability as total deviation in allocated reward mass.

\paragraph{Empirical Observations.}
Figure~\ref{fig:reward-fairness-dataset} reports reward fairness as a function of the non-IID parameter $\alpha$.
When data distributions are close to IID (large $\alpha$), all schemes achieve similar fairness levels, and uniform allocation is often sufficient.
In contrast, under strong non-IID conditions (small $\alpha$), fairness degrades substantially across all schemes.
While confidence- and agreement-based allocations can outperform uniform allocation in some settings, they may also be less fair than uniform allocation under severe label skew.

Figure~\ref{fig:reward-fairness-non-iid} visualizes the relationship between ideal merit share $m_k$ and allocated reward share $r_k$.
The diagonal line corresponds to perfect merit alignment.
For $K=5$ clients, perfectly fair allocation concentrates points near $m_k \approx 0.2$, whereas dispersion away from the diagonal indicates misalignment.
Under strong non-IID conditions, agreement- and confidence-based schemes exhibit substantial deviation from the diagonal, whereas all schemes converge toward merit-aligned behavior as $\alpha$ increases.

\paragraph{Discussion.}
These results highlight the intrinsic difficulty of designing fair incentive mechanisms for FI without labels.
In highly non-IID settings, clients may contribute models of vastly different quality, yet label-free reward schemes do not consistently prioritize high-merit contributors and may even be less fair than uniform allocation.
Moreover, these findings raise broader questions about whether individual model accuracy alone is an appropriate proxy for contribution, or whether reward allocation should instead reflect each model’s influence on the ensemble decision.
We view reward fairness in label-free FI as a fundamental open problem.

\section{Discussion: Incentive Design under Privacy and Observability Constraints}
\label{subsec:incentive_design}

Designing fair and sustainable incentive mechanisms remains a fundamental challenge in FI.
Unlike FL, where contribution can often be verified through observable training signals such as gradients or model updates, FI operates under strict privacy constraints that fundamentally limit observability.
Inference execution is typically performed off-chain and within protected environments (e.g., SMPC), rendering both intermediate computations and final outputs opaque to external verifiers.

While blockchain-based smart contracts provide transparency and tamper-resistant settlement, this transparency can conflict with privacy objectives.
In particular, differentiated rewards may implicitly reveal information about individual model behavior or specialization.
Over repeated inference queries, such reward signals can accumulate and enable adversaries to infer sensitive properties of local training distributions or model capabilities.
This tension highlights an inherent trade-off between incentive expressiveness and privacy preservation.

Beyond contribution measurement, FI introduces a more fundamental challenge: \emph{verifiable completion of inference}.
Because inference is executed privately and off-chain, a smart contract cannot directly verify whether inference was correctly performed, whether the client received the result, or whether participating model providers followed the prescribed protocol.
This lack of observability creates a structural asymmetry in reward settlement.

A naïve approach is to rely on the client to trigger reward distribution after receiving the inference result.
However, such a design enables strategic behavior, as a rational client may withhold the final on-chain call after obtaining the prediction, preventing reward allocation.
Conversely, allowing model providers to trigger settlement introduces the opposite risk, whereby rewards could be claimed without verifiable evidence that inference was executed or that the client successfully received the output.
Introducing a trusted third party to resolve this asymmetry undermines the privacy and decentralization goals of FI.

In FedSEI, we intentionally adopt a minimal, trust-assumed settlement logic to isolate the interaction between privacy-preserving inference, collaborative utility, and incentive ambiguity.
This design choice allows us to empirically study incentive behavior without introducing additional cryptographic assumptions or protocol complexity.
However, it also exposes a critical open problem: how to securely and fairly trigger reward settlement when inference execution itself is unobservable.

One promising approach is to implement \emph{refundable collateral mechanisms} to address the ``finalization problem,'' where a client, having already received the inference result off-chain, lacks a rational incentive to trigger on-chain settlement.
In standard escrow designs, the client may simply withhold the final confirmation to avoid additional gas costs.
To counter this, the system can require the client to deposit a collateral exceeding the service fee (e.g., fee + stake), where the surplus stake is refunded \emph{only} upon the successful verification of job completion.
This design shifts the trust assumption from protocol correctness to economic rationality, compelling the client to finalize the transaction to recover their collateral.

More generally, designing lightweight and privacy-preserving proofs of inference completion remains an open research problem.
Potential avenues include verifiable MPC~\cite{simunic_verifiable_2021}, zero-knowledge attestations of protocol execution~\cite{gabizon_plonk_2019, south_verifiable_2024}, or hybrid on-chain/off-chain signaling mechanisms~\cite{chan_optimistic_2025}.
These approaches must balance verifiability with privacy, avoiding additional leakage about inputs, outputs, or model internals.

Taken together, these challenges suggest that incentive design in FI is not a straightforward extension of existing blockchain-based reward mechanisms.
Rather, it constitutes a distinct systems problem shaped by the interplay between privacy, limited observability, and strategic participant behavior.
Addressing this problem is essential for building FI systems that are not only technically correct, but also economically sustainable in real-world deployments.

\section{Conclusion}

This work examined FI as a collaborative paradigm operating strictly at inference time, where independently trained and privately owned models cooperate without sharing data or model parameters.
We argued that FI is fundamentally governed by two requirements: inference-time privacy preservation and measurable performance gains through collaboration.

We instantiated this perspective through \emph{FedSEI}, which combines SMPC for confidentiality and ensemble-based aggregation for collaborative utility.
Our system-level evaluation characterized the inherent overhead of privacy-preserving inference and demonstrated that ensemble effectiveness under non-IID data is highly context-dependent.
While uniform aggregation provides a strong baseline in many settings, it does not universally guarantee performance improvements, particularly under severe data heterogeneity.

These findings indicate that FI constitutes a distinct design space that cannot be directly reduced to FL or conventional ensemble methods.
Effective collaborative inference requires joint consideration of privacy enforcement, aggregation behavior, and deployment constraints.

Beyond empirical findings, this work contributes by explicitly articulating the incentive ambiguity inherent to federated inference without labels.
By demonstrating that label-free reward schemes may fail to consistently prioritize high-merit contributors—and can even be less fair than uniform allocation under severe non-IID conditions—we clarify that the challenge is not merely algorithmic, but structural.
We hope this work provides a clear foundation for future research on principled and practical FI systems.

\section{Open Problems and Research Directions}
\label{sec:open_problems}

This section outlines open research problems motivated by our empirical findings.
Rather than incremental extensions, we focus on structural challenges that arise when privacy, collaboration, and incentives are enforced at inference time.

\subsection{Improving the Efficiency of Privacy-Preserving Inference}

Our results show that SMPC-based inference incurs substantial computational and communication overhead, which often dominates end-to-end latency~\cite{keller2020mp,rathee2020cryptflow2,huang2022cheetah}.
This highlights several open questions in the design of efficient privacy-preserving inference pipelines. First, the current practice of directly mapping plaintext neural network operations to generic SMPC protocols may be suboptimal. Exploring SMPC-aware model architectures or protocol-level optimizations for common operations, such as non-linear activations, remains an open direction~\cite{liu2017oblivious,rouhani2018deepsecure,mishra2020delphi}. Second, it is unclear whether protecting all model components under SMPC is always necessary.
Hybrid designs~\cite{jarin2021pricure,zhao2025fedinf} that selectively protect sensitive layers while executing others in plaintext may offer improved efficiency under acceptable privacy guarantees. Third, combining SMPC with other privacy-preserving primitives, such as HE or TEEs, presents a promising direction.
Understanding the trade-offs between security assumptions, performance, and deployability remains an open problem~\cite{tramer2018slalom,juvekar2018gazelle}. Finally, wide-area deployments emphasize the impact of network topology on secure inference.
Topology-aware execution or hierarchical protocol designs may mitigate latency~\cite{liu2020client}, but introduce new trust and attack considerations that require careful analysis.

\subsection{Beyond Ensemble-Based Collaboration}

While ensemble inference offers a simple and model-agnostic collaboration mechanism, our results indicate that it does not universally yield performance gains under severe non-IID data heterogeneity.
This motivates the exploration of collaboration strategies beyond ensemble aggregation. One direction is model fusion~\cite{wortsman2022model}, which merges independently trained models into a single representation.
Although fusion may reduce diversity and impose architectural constraints, it could offer computational advantages and different robustness properties. Another direction is routing-based collaboration inspired by mixture-of-experts architectures~\cite{zhou2022mixture}.
Instead of aggregating all models, routing mechanisms could selectively activate a subset of models per query.
A key challenge is learning such routing policies without access to raw data or ground-truth labels, relying only on unlabeled queries or privacy-preserving signals~\cite{farhat2025learning,zhou2025cryptomoe}. More broadly, collaboration mechanisms may benefit from limited proxy information. Even weak signals, such as small amounts of public or synthetic data, could enable more adaptive collaboration while preserving the privacy assumptions of FI~\cite{pmlr-v139-zhu21b}.

\subsection{Incentive Design under Label-Free Collaborative Inference}

While this work incorporates a minimal on-chain incentive mechanism for service settlement, our empirical findings highlight that incentive design remains a fundamental open problem in FI.
Unlike training-based collaboration, inference-time collaboration typically lacks access to ground-truth labels, making it difficult to objectively measure individual model contributions~\cite{zhan2021survey,liu2022gtg}. In the absence of labels, commonly used proxies such as prediction confidence, agreement rates, or stability under perturbation provide only indirect and often unreliable signals of contribution. As demonstrated by our ensemble experiments, the relative utility of a model is highly context-dependent and can vary significantly across queries, datasets, and degrees of data heterogeneity.
This variability complicates the design of reward functions that are both fair and manipulation-resistant~\cite{borovac2022ensemble}. Moreover, incentive mechanisms must operate under the same privacy constraints as inference itself. Direct evaluation of model performance or contribution may conflict with confidentiality requirements, limiting the information available for reward allocation. Balancing economic incentives with privacy preservation, therefore, introduces an inherent tension in FI systems.

Addressing these challenges requires new approaches to contribution estimation that are compatible with label-free, privacy-preserving settings.
Whether through richer proxy signals, limited public references, or novel cryptographic primitives, principled incentive design remains a key open research direction for practical FI.

\subsection{Federated Inference in the Era of Large Language Models}

The emergence of large language models (LLMs) raises new questions about the role of FI.
Direct privacy-preserving execution of LLMs remains challenging due to their computational scale~\cite{ICLR2025_6715b4e9,deencryptedllm}. However, FI may play a complementary role at higher levels of the inference pipeline.
For example, FI-style collaboration could be applied to tools, agents, or decision modules orchestrated by LLMs, rather than within the core model itself~\cite{schick2023toolformer,yu2025survey}.
In such settings, inference latency constraints are often less stringent. Moreover, emerging AI workflows increasingly tolerate longer execution times than traditional real-time services.
This shift may expand the practical deployment space of FI, particularly in batch-oriented or decision-critical applications.
Clarifying the role of FI in these evolving systems remains an open research direction.

\printcredits

\section*{Data availability}
The datasets used in this study are publicly available.

\section*{Declaration of competing interest}
The authors declare that they have no known competing financial interests or personal relationships that could have appeared to influence the work reported in this paper.

\section*{Acknowledgments}
This work was supported by the research centre NCS2030 (RCN project number 331644). The authors acknowledge the collaboration with the centre.

\appendix
\section*{Appendix}
\section{Details of Weight Allocation Strategies}
\label{app:weighting_details}

\begin{table}[h]
\centering
\caption{Computation overhead of secure ensemble methods (in ms) measured under a single-host multi-processing setup, using two input samples (batch size = 2). Reported values are averaged over three runs.}

\label{tab:ensemble-ms-stats}
\begin{tabular}{lr}
\hline
\textbf{Ensemble Method} & \textbf{Mean $\pm$ Std Dev (ms)} \\ \hline
Hard Voting              & $417.3 \pm 24.7$                \\
Uniform Soft Voting              & $174.3 \pm 34.4$                \\
Entropy-based Weighted Voting           & $371.3 \pm 31.9$                \\
Spectral-based Weighted Voting          & $513.7 \pm 9.8$                 \\
TTA (L2)-based Weighted Voting          & $2441.3 \pm 58.6$               \\ \hline
\end{tabular}
\end{table}

In this appendix, we detail the specific instantiation of the aggregation weights
$\{w_i\}_{i=1}^{N}$ introduced in Section~\ref{sec:inst-FedSEI}.

Throughout this appendix, we assume a generic ensemble prediction of the form
\begin{equation}
y = \sum_{i=1}^{N} w_i(x) \cdot \mathbf{p}_i(x),
\end{equation}
where $\mathbf{p}_i(x)$ denotes the predictive distribution produced by model $M_i$ for input $x$.
The following subsections focus exclusively on how the aggregation weights $w_i$ are determined
under different design choices.
The computational overheads incurred by each ensemble strategy are summarized in
Table~\ref{tab:ensemble-ms-stats}.

\subsection{Entropy-Based Weighting}
\label{app:entropy_weighting}

Entropy-based weighting assigns dynamic, input-dependent aggregation weights based on
the predictive uncertainty of each model.

For a query input $x$, let $\mathbf{z}_{i} \in \mathbb{R}^C$ denote the output logits of model $M_i$,
and let $\mathbf{p}_{i} = \mathrm{softmax}(\mathbf{z}_{i})$ be the corresponding predictive distribution.
We quantify uncertainty using Shannon entropy:
\begin{equation}
H_i(x) = - \sum_{c=1}^{C} p_i^{(c)} \log p_i^{(c)}.
\end{equation}

Models with lower entropy are interpreted as being more confident.
Accordingly, aggregation weights are defined via a softmax over negative entropy values:
\begin{equation}
w_i(x) = \frac{\exp(-\beta \cdot H_i(x))}{\sum_{j=1}^{N} \exp(-\beta \cdot H_j(x))},
\end{equation}
where $\beta > 0$ controls the sharpness of the weighting. Unless otherwise stated, we set $\beta = 1$ in all experiments.

\subsection{Spectral Weighting (Global Static Weights)}
\label{app:spectral_weighting}

Spectral weighting estimates sample-invariant aggregation weights based on the global
consistency of model confidence patterns across a shared calibration dataset.
This method relies on aggregating confidence statistics over a sufficiently large set of
unlabeled inputs to estimate relative model reliability.

Let $\mathcal{D}_{\text{cal}} = \{x^{(1)}, \dots, x^{(S)}\}$ be an unlabeled calibration set.
For model $M_i$ and sample $x^{(s)}$, let
$\mathbf{p}_{i,s} = \mathrm{softmax}(\mathbf{z}_{i,s})$ denote the predictive distribution.

We define the confidence score
\begin{equation}
\phi_{i,s} = \max_{c} \, p_{i,s}^{(c)},
\end{equation}
and center it per model:
\begin{equation}
\tilde{\phi}_{i,s} = \phi_{i,s} - \frac{1}{S} \sum_{j=1}^{S} \phi_{i,j}.
\end{equation}

Let $\tilde{\mathbf{\Phi}} \in \mathbb{R}^{N \times S}$ collect the centered scores.
The correlation matrix is defined as
\begin{equation}
\mathbf{C} = \frac{1}{S-1} \tilde{\mathbf{\Phi}} \tilde{\mathbf{\Phi}}^\top.
\end{equation}

Let $\mathbf{v}$ be the principal eigenvector of $\mathbf{C}$.
The global aggregation weights are then given by
\begin{equation}
w_i = \frac{|\mathbf{v}_i|}{\sum_{j=1}^{N} |\mathbf{v}_j|}.
\end{equation}

In our experiments, spectral weighting is applied over large evaluation sets, where
sufficient samples are available to yield stable estimates of confidence correlations.
However, in practical deployment scenarios where inference is performed on a single or
very small number of queries, this approach becomes ill-defined, as the required
second-order statistics cannot be reliably estimated.
Accordingly, spectral weighting should be interpreted as a batch-level aggregation
mechanism rather than a per-query inference strategy.

\subsection{TTA-Based Instability Weighting}
\label{app:tta_weighting}

TTA-based weighting assigns input-dependent aggregation weights based on prediction
instability under test-time augmentations.

For a query input $x$, we generate $T$ augmented views $\{x_{(t)}\}_{t=1}^{T}$.
In our experiments, augmentations consist of random rotations within $\pm10^\circ$, and
we set $T=2$ to limit the additional inference cost, as larger $T$ increases latency
linearly.
Model $M_i$ produces logits $\mathbf{z}_{i}^{(t)}$ for each augmented view.
Prediction instability is measured as
\begin{equation}
d_{i}(x) = \frac{1}{T} \sum_{t=1}^{T}
\left\| \mathbf{z}_{i}^{(t)} - \mathbf{z}_{i}^{(0)} \right\|_2.
\end{equation}

Aggregation weights are then computed as
\begin{equation}
w_{i}(x) = \frac{\exp(\gamma \cdot d_{i}(x))}{\sum_{j=1}^{N} \exp(\gamma \cdot d_{j}(x))},
\end{equation}
where $\gamma > 0$ controls the sharpness of weighting.
Unless otherwise stated, we set $\gamma = 1.0$.

Unlike conventional interpretations that associate lower instability with higher reliability,
we intentionally assign higher weights to models exhibiting larger prediction variation.
In FI settings with severe label skew, models can appear overly stable on inputs associated
with unseen labels, often producing confident but incorrect predictions.
In such cases, controlled sensitivity to test-time perturbations can serve as a proxy signal
for a model’s ability to generalize beyond its local label support.

\bibliographystyle{unsrt}

\bibliography{main}

@article{knott2021crypten,
  title={Crypten: Secure multi-party computation meets machine learning},
  author={Knott, Brian and Venkataraman, Shobha and Hannun, Awni and Sengupta, Shubho and Ibrahim, Mark and van der Maaten, Laurens},
  journal={Advances in Neural Information Processing Systems},
  volume={34},
  pages={4961--4973},
  year={2021}
}

@inproceedings{keller2020mp,
  title={MP-SPDZ: A versatile framework for multi-party computation},
  author={Keller, Marcel},
  booktitle={Proceedings of the 2020 ACM SIGSAC conference on computer and communications security},
  pages={1575--1590},
  year={2020}
}

@article{RONG2025100277,
title = {Federated Large Domain Model System},
journal = {Blockchain: Research and Applications},
volume = {6},
number = {3},
pages = {100277},
year = {2025},
issn = {2096-7209},
doi = {https://doi.org/10.1016/j.bcra.2025.100277},
url = {https://www.sciencedirect.com/science/article/pii/S2096720925000041},
author = {Chunming Rong and Jungwon Seo and Zihan Zhao and Ferhat Ozgur Catak and Jiahui Geng and Martin Gilje Jaatun},

}

@article{kuo2025research,
  title={Research in Collaborative Learning Does Not Serve Cross-Silo Federated Learning in Practice},
  author={Kuo, Kevin and Yadav, Chhavi and Smith, Virginia},
  journal={arXiv preprint arXiv:2510.12595},
  year={2025}
}

@inproceedings{mcmahan2017communication,
  title={Communication-efficient learning of deep networks from decentralized data},
  author={McMahan, Brendan and Moore, Eider and Ramage, Daniel and Hampson, Seth and y Arcas, Blaise Aguera},
  booktitle={Artificial intelligence and statistics},
  pages={1273--1282},
  year={2017},
  organization={PMLR}
}

@article{ahsen2019unsupervised,
  title={Unsupervised evaluation and weighted aggregation of ranked classification predictions},
  author={Ahsen, Mehmet Eren and Vogel, Robert M and Stolovitzky, Gustavo A},
  journal={Journal of Machine Learning Research},
  volume={20},
  number={166},
  pages={1--40},
  year={2019}
}

@inproceedings{ayhan2018test,
  title={Test-time data augmentation for estimation of heteroscedastic aleatoric uncertainty in deep neural networks},
  author={Ayhan, Murat Seckin and Berens, Philipp},
  booktitle={Medical Imaging with Deep Learning},
  year={2018}
}

@inproceedings{jarin2021pricure,
  title={PRICURE: Privacy-preserving collaborative inference in a multi-party setting},
  author={Jarin, Ismat and Eshete, Birhanu},
  booktitle={Proceedings of the 2021 ACM workshop on security and privacy analytics},
  pages={25--35},
  year={2021}
}

@article{kairouz2021advances,
  title={Advances and open problems in federated learning},
  author={Kairouz, Peter and McMahan, H Brendan and Avent, Brendan and Bellet, Aur{\'e}lien and Bennis, Mehdi and Bhagoji, Arjun Nitin and Bonawitz, Kallista and Charles, Zachary and Cormode, Graham and Cummings, Rachel and others},
  journal={Foundations and trends{\textregistered} in machine learning},
  volume={14},
  number={1--2},
  pages={1--210},
  year={2021},
  publisher={Now Publishers, Inc.}
}

@article{lecun2002gradient,
  title={Gradient-based learning applied to document recognition},
  author={LeCun, Yann and Bottou, L{\'e}on and Bengio, Yoshua and Haffner, Patrick},
  journal={Proceedings of the IEEE},
  volume={86},
  number={11},
  pages={2278--2324},
  year={2002},
  publisher={Ieee}
}

@inproceedings{he2016deep,
  title={Deep residual learning for image recognition},
  author={He, Kaiming and Zhang, Xiangyu and Ren, Shaoqing and Sun, Jian},
  booktitle={Proceedings of the IEEE conference on computer vision and pattern recognition},
  pages={770--778},
  year={2016}
}

@inproceedings{juvekar2018gazelle,
  title={$\{$GAZELLE$\}$: A low latency framework for secure neural network inference},
  author={Juvekar, Chiraag and Vaikuntanathan, Vinod and Chandrakasan, Anantha},
  booktitle={27th USENIX security symposium (USENIX security 18)},
  pages={1651--1669},
  year={2018}
}

@inproceedings{
reddi2021adaptive,
title={Adaptive Federated Optimization},
author={Sashank J. Reddi and Zachary Charles and Manzil Zaheer and Zachary Garrett and Keith Rush and Jakub Kone{\v{c}}n{\'y} and Sanjiv Kumar and Hugh Brendan McMahan},
booktitle={International Conference on Learning Representations},
year={2021},
url={https://openreview.net/forum?id=LkFG3lB13U5}
}

@inproceedings{
oh2022fedbabu,
title={Fed{BABU}: Toward Enhanced Representation for Federated Image Classification},
author={Jaehoon Oh and SangMook Kim and Se-Young Yun},
booktitle={International Conference on Learning Representations},
year={2022},
url={https://openreview.net/forum?id=HuaYQfggn5u}
}

@inproceedings{nguyen2022federated,
  title={Federated learning with buffered asynchronous aggregation},
  author={Nguyen, John and Malik, Kshitiz and Zhan, Hongyuan and Yousefpour, Ashkan and Rabbat, Mike and Malek, Mani and Huba, Dzmitry},
  booktitle={International conference on artificial intelligence and statistics},
  pages={3581--3607},
  year={2022},
  organization={PMLR}
}

@inproceedings{haddadpour2021federated,
  title={Federated learning with compression: Unified analysis and sharp guarantees},
  author={Haddadpour, Farzin and Kamani, Mohammad Mahdi and Mokhtari, Aryan and Mahdavi, Mehrdad},
  booktitle={International Conference on Artificial Intelligence and Statistics},
  pages={2350--2358},
  year={2021},
  organization={PMLR}
}

@inproceedings{bonawitz2017practical,
  title={Practical secure aggregation for privacy-preserving machine learning},
  author={Bonawitz, Keith and Ivanov, Vladimir and Kreuter, Ben and Marcedone, Antonio and McMahan, H Brendan and Patel, Sarvar and Ramage, Daniel and Segal, Aaron and Seth, Karn},
  booktitle={proceedings of the 2017 ACM SIGSAC Conference on Computer and Communications Security},
  pages={1175--1191},
  year={2017}
}

@article{wei2020federated,
  title={Federated learning with differential privacy: Algorithms and performance analysis},
  author={Wei, Kang and Li, Jun and Ding, Ming and Ma, Chuan and Yang, Howard H and Farokhi, Farhad and Jin, Shi and Quek, Tony QS and Poor, H Vincent},
  journal={IEEE transactions on information forensics and security},
  volume={15},
  pages={3454--3469},
  year={2020},
  publisher={IEEE}
}

@article{kumar2023impact,
  title={The impact of adversarial attacks on federated learning: A survey},
  author={Kumar, Kummari Naveen and Mohan, Chalavadi Krishna and Cenkeramaddi, Linga Reddy},
  journal={IEEE Transactions on Pattern Analysis and Machine Intelligence},
  volume={46},
  number={5},
  pages={2672--2691},
  year={2023},
  publisher={IEEE}
}

@article{buterin2014next,
  title={A next-generation smart contract and decentralized application platform},
  author={Buterin, Vitalik and others},
  journal={white paper},
  volume={3},
  number={37},
  pages={2--1},
  year={2014}
}

@inproceedings{wortsman2022model,
  title={Model soups: averaging weights of multiple fine-tuned models improves accuracy without increasing inference time},
  author={Wortsman, Mitchell and Ilharco, Gabriel and Gadre, Samir Ya and Roelofs, Rebecca and Gontijo-Lopes, Raphael and Morcos, Ari S and Namkoong, Hongseok and Farhadi, Ali and Carmon, Yair and Kornblith, Simon and others},
  booktitle={International conference on machine learning},
  pages={23965--23998},
  year={2022},
  organization={PMLR}
}

@article{zhou2022mixture,
  title={Mixture-of-experts with expert choice routing},
  author={Zhou, Yanqi and Lei, Tao and Liu, Hanxiao and Du, Nan and Huang, Yanping and Zhao, Vincent and Dai, Andrew M and Le, Quoc V and Laudon, James and others},
  journal={Advances in Neural Information Processing Systems},
  volume={35},
  pages={7103--7114},
  year={2022}
}

@inproceedings{ding2021incentive,
  title={Incentive mechanism design for distributed coded machine learning},
  author={Ding, Ningning and Fang, Zhixuan and Duan, Lingjie and Huang, Jianwei},
  booktitle={IEEE INFOCOM 2021-IEEE Conference on Computer Communications},
  pages={1--10},
  year={2021},
  organization={IEEE}
}

@inproceedings{gentry2009fully,
  title={Fully homomorphic encryption using ideal lattices},
  author={Gentry, Craig},
  booktitle={Proceedings of the forty-first annual ACM symposium on Theory of computing},
  pages={169--178},
  year={2009}
}

@article{sagar2021confidential,
  title={Confidential machine learning on untrusted platforms: a survey},
  author={Sagar, Sharma and Keke, Chen},
  journal={Cybersecurity},
  volume={4},
  number={1},
  pages={30},
  year={2021},
  publisher={Springer}
}

@article{chen2025privacy,
  title={Privacy-preserving machine learning based on cryptography: a survey},
  author={Chen, Congcong and Wei, Lifei and Xie, Jintao and Shi, Yang},
  journal={ACM Transactions on Knowledge Discovery from Data},
  volume={19},
  number={4},
  pages={1--33},
  year={2025},
  publisher={ACM New York, NY}
}

@article{seo2026gc,
  title={GC-Fed: Gradient Centralized Federated Learning with Partial Client Participation},
  author={Seo, Jungwon and Catak, Ferhat Ozgur and Rong, Chunming and Hong, Kibeom and Kim, Minhoe},
  journal={Information Fusion},
  pages={104148},
  year={2026},
  publisher={Elsevier}
}

@inproceedings{medmnistv1,
    title={MedMNIST Classification Decathlon: A Lightweight AutoML Benchmark for Medical Image Analysis},
    author={Yang, Jiancheng and Shi, Rui and Ni, Bingbing},
    booktitle={IEEE 18th International Symposium on Biomedical Imaging (ISBI)},
    pages={191--195},
    year={2021}
}

@article{krizhevsky2009learning,
  title={Learning multiple layers of features from tiny images},
  author={Krizhevsky, Alex and Hinton, Geoffrey and others},
  year={2009},
  publisher={Toronto, ON, Canada}
}

@article{xiao2017fashion,
  title={Fashion-mnist: a novel image dataset for benchmarking machine learning algorithms},
  author={Xiao, Han and Rasul, Kashif and Vollgraf, Roland},
  journal={arXiv preprint arXiv:1708.07747},
  year={2017}
}

@inproceedings{cohen2017emnist,
  title={EMNIST: Extending MNIST to handwritten letters},
  author={Cohen, Gregory and Afshar, Saeed and Tapson, Jonathan and Van Schaik, Andre},
  booktitle={2017 international joint conference on neural networks (IJCNN)},
  pages={2921--2926},
  year={2017},
  organization={IEEE}
}

@inproceedings{yuan_mlink_2022,
	title = {{MLink}: {Linking} {Black}-{Box} {Models} for {Collaborative} {Multi}-{Model} {Inference}},
	volume = {36},
	copyright = {Copyright (c) 2022 Association for the Advancement of Artificial Intelligence},
	issn = {2374-3468},
	shorttitle = {{MLink}},
	url = {https://ojs.aaai.org/index.php/AAAI/article/view/21180},
	doi = {10.1609/aaai.v36i9.21180},
	language = {en},
	number = {9},
	urldate = {2026-01-28},
	journal = {Proceedings of the AAAI Conference on Artificial Intelligence},
	author = {Yuan, Mu and Zhang, Lan and Li, Xiang-Yang},
	month = jun,
	year = {2022},
	pages = {9475--9483},
}

@article{zhao2025fedinf,
  title={{FEDINF}: An Efficient and Secure Inference with Federated Participants},
  author={Zhao, Bowen and Guo, Weibin and Chen, Jiahui and Xiao, Yang and Zhai, Liang and Pei, Qingqi},
  journal={IEEE Transactions on Computers},
  year={2025},
  publisher={IEEE},
  doi={10.1109/TC.2025.3642408}
}

@inproceedings{fan2025federated,
  title={Federated Inference: Towards Collaborative and Privacy-Preserving Inference Over Edge Devices},
  author={Fan, B. and Su, X. and Tarkoma, S. and Hui, P.},
  booktitle={Proceedings of the ACM SIGCOMM 2025 Posters and Demos},
  year={2025}
}

@article{zhou2025towards,
  title={Towards Federated Inference: An Online Model Ensemble Framework for Cooperative Edge AI},
  author={Zhou, Z. and Xie, J. and Huang, M. and Ouyang, T. and others},
  journal={IEEE INFOCOM 2025 Conference on Computer Communications},
  year={2025},
  publisher={IEEE}
}

@inproceedings{tanikanti2025first,
  title={{FIRST}: Federated Inference Resource Scheduling Toolkit for Scientific AI Model Access},
  author={Tanikanti, A. and Côté, B. and Guo, Y. and Chen, L. and Saint-Cyr, N. and others},
  booktitle={Proceedings of the International Conference for High Performance Computing, Networking, Storage and Analysis (SC '25)},
  year={2025}
}

@article{dhasade2025robust,
  title={Robust Federated Inference},
  author={Dhasade, Akash and Farhadkhani, Sadegh and Guerraoui, Rachid and Gupta, Nirupam and Jacovella, Maxime and Kermarrec, Anne-Marie and Pinot, Rafael},
  journal={arXiv preprint arXiv:2510.00310},
  year={2025}
}

@inproceedings{alipanahloo2025rzkfl,
  title={{RzkFL}: A Verifiable, Fast and Privacy-Preserving Framework for Federated Learning Inference Using Recursive Zero-Knowledge Proofs},
  author={Alipanahloo, Z. and Duchesne, M. and others},
  booktitle={2025 IEEE International Conference on Blockchain and Cryptocurrency (ICBC)},
  year={2025}
}

@article{zhu2024federated,
  title={Federated Inference With Reliable Uncertainty Quantification Over Wireless Channels via Conformal Prediction},
  author={Zhu, M. and Zecchin, M. and Park, S. and Guo, C. and others},
  journal={IEEE Transactions on Wireless Communications},
  year={2024},
  publisher={IEEE}
}

@article{jonker2025bayesian,
  title={Bayesian Federated Inference for regression models based on non-shared medical center data},
  author={Jonker, M. A. and Pazira, H. and Coolen, A. C. C.},
  journal={Research Synthesis Methods},
  year={2025},
  publisher={Wiley}
}

@article{vo2025federated,
  title={Federated causal inference from observational data},
  author={Vo, T. V. and Lee, Y. and Leong, T. Y.},
  journal={Machine Learning},
  year={2025},
  publisher={Springer}
}

@article{friston2024federated,
  title={Federated inference and belief sharing},
  author={Friston, K. J. and Parr, T. and Heins, C. and Constant, A. and others},
  journal={Neuroscience \& Biobehavioral Reviews},
  year={2024},
  publisher={Elsevier}
}

@article{simunic_verifiable_2021,
	title = {Verifiable computing applications in blockchain},
	volume = {9},
	url = {https://ieeexplore.ieee.org/abstract/document/9620108/},
	urldate = {2026-02-01},
	journal = {IEEE access},
	author = {Šimunić, Silvio and Bernaca, Dalen and Lenac, Kristijan},
	year = {2021},
	note = {Publisher: IEEE},
	pages = {156729--156745},
}

@misc{south_verifiable_2024,
	title = {Verifiable evaluations of machine learning models using {zkSNARKs}},
	url = {http://arxiv.org/abs/2402.02675},
	doi = {10.48550/arXiv.2402.02675},
	urldate = {2026-02-01},
	publisher = {arXiv},
	author = {South, Tobin and Camuto, Alexander and Jain, Shrey and Nguyen, Shayla and Mahari, Robert and Paquin, Christian and Morton, Jason and Pentland, Alex 'Sandy'},
	month = may,
	year = {2024},
	note = {arXiv:2402.02675 [cs]},
}

@article{gabizon_plonk_2019,
	title = {Plonk: {Permutations} over lagrange-bases for oecumenical noninteractive arguments of knowledge},
	shorttitle = {Plonk},
	url = {https://eprint.iacr.org/2019/953},
	urldate = {2026-02-01},
	journal = {Cryptology ePrint Archive},
	author = {Gabizon, Ariel and Williamson, Zachary J. and Ciobotaru, Oana},
	year = {2019},
}

@misc{chan_optimistic_2025,
	title = {Optimistic {TEE}-rollups: {A} hybrid architecture for scalable and verifiable generative {AI} inference on blockchain},
	shorttitle = {Optimistic {TEE}-{Rollups}},
	url = {http://arxiv.org/abs/2512.20176},
	doi = {10.48550/arXiv.2512.20176},
	urldate = {2026-02-01},
	publisher = {arXiv},
	author = {Chan, Aaron and Ding, Alex and Chen, Frank and Wu, Alan and Zhang, Bruce and Tian, Arther},
	month = dec,
	year = {2025},
	note = {arXiv:2512.20176 [cs]},
}

@article{seo2025understanding,
  title={Understanding federated learning from iid to non-iid dataset: An experimental study},
  author={Seo, Jungwon and Catak, Ferhat Ozgur and Rong, Chunming},
  journal={arXiv preprint arXiv:2502.00182},
  year={2025}
}

@misc{li2019impact,
  title={The impact of GDPR on global technology development},
  author={Li, He and Yu, Lu and He, Wu},
  journal={Journal of Global Information Technology Management},
  volume={22},
  number={1},
  pages={1--6},
  year={2019},
  publisher={Taylor \& Francis}
}

@article{mothukuri2021survey,
  title={A survey on security and privacy of federated learning},
  author={Mothukuri, Viraaji and Parizi, Reza M and Pouriyeh, Seyedamin and Huang, Yan and Dehghantanha, Ali and Srivastava, Gautam},
  journal={Future Generation Computer Systems},
  volume={115},
  pages={619--640},
  year={2021},
  publisher={Elsevier}
}

@inproceedings{philipp2020machine,
  title={Machine learning as a service: Challenges in research and applications},
  author={Philipp, Robert and Mladenow, Andreas and Strauss, Christine and V{\"o}lz, Alexander},
  booktitle={Proceedings of the 22nd International Conference on Information Integration and Web-based Applications \& Services},
  pages={396--406},
  year={2020}
}

@inproceedings{gan2023model,
  title={Model-as-a-service (MaaS): A survey},
  author={Gan, Wensheng and Wan, Shicheng and Philip, S Yu},
  booktitle={2023 IEEE International Conference on Big Data (BigData)},
  pages={4636--4645},
  year={2023},
  organization={IEEE}
}

@article{li2025collaborative,
  title={Collaborative inference and learning between edge slms and cloud LLMs: A survey of algorithms, execution, and open challenges},
  author={Li, Senyao and Wang, Haozhao and Xu, Wenchao and Zhang, Rui and Guo, Song and Yuan, Jingling and Zhong, Xian and Zhang, Tianwei and Li, Ruixuan},
  journal={arXiv preprint arXiv:2507.16731},
  year={2025}
}

@article{zhang2022security,
  title={Security and privacy threats to federated learning: Issues, methods, and challenges},
  author={Zhang, Junpeng and Zhu, Hui and Wang, Fengwei and Zhao, Jiaqi and Xu, Qi and Li, Hui},
  journal={Security and Communication Networks},
  volume={2022},
  number={1},
  pages={2886795},
  year={2022},
  publisher={Wiley Online Library}
}

@inproceedings{shokri2017membership,
  title={Membership inference attacks against machine learning models},
  author={Shokri, Reza and Stronati, Marco and Song, Congzheng and Shmatikov, Vitaly},
  booktitle={2017 IEEE symposium on security and privacy (SP)},
  pages={3--18},
  year={2017},
  organization={IEEE}
}

@inproceedings{androulaki2018hyperledger,
  title={Hyperledger fabric: a distributed operating system for permissioned blockchains},
  author={Androulaki, Elli and Barger, Artem and Bortnikov, Vita and Cachin, Christian and Christidis, Konstantinos and De Caro, Angelo and Enyeart, David and Ferris, Christopher and Laventman, Gennady and Manevich, Yacov and others},
  booktitle={Proceedings of the thirteenth EuroSys conference},
  pages={1--15},
  year={2018}
}

@inproceedings{rathee2020cryptflow2,
  title={Cryptflow2: Practical 2-party secure inference},
  author={Rathee, Deevashwer and Rathee, Mayank and Kumar, Nishant and Chandran, Nishanth and Gupta, Divya and Rastogi, Aseem and Sharma, Rahul},
  booktitle={Proceedings of the 2020 ACM SIGSAC conference on computer and communications security},
  pages={325--342},
  year={2020}
}

@inproceedings{huang2022cheetah,
  title={Cheetah: Lean and fast secure $\{$Two-Party$\}$ deep neural network inference},
  author={Huang, Zhicong and Lu, Wen-jie and Hong, Cheng and Ding, Jiansheng},
  booktitle={31st USENIX Security Symposium (USENIX Security 22)},
  pages={809--826},
  year={2022}
}

@inproceedings{liu2017oblivious,
  title={Oblivious neural network predictions via minionn transformations},
  author={Liu, Jian and Juuti, Mika and Lu, Yao and Asokan, Nadarajah},
  booktitle={Proceedings of the 2017 ACM SIGSAC conference on computer and communications security},
  pages={619--631},
  year={2017}
}

@inproceedings{rouhani2018deepsecure,
  title={Deepsecure: Scalable provably-secure deep learning},
  author={Rouhani, Bita Darvish and Riazi, M Sadegh and Koushanfar, Farinaz},
  booktitle={Proceedings of the 55th annual design automation conference},
  pages={1--6},
  year={2018}
}

@inproceedings{mishra2020delphi,
  title={Delphi: A cryptographic inference system for neural networks},
  author={Mishra, Pratyush and Lehmkuhl, Ryan and Srinivasan, Akshayaram and Zheng, Wenting and Popa, Raluca Ada},
  booktitle={Proceedings of the 2020 Workshop on Privacy-Preserving Machine Learning in Practice},
  pages={27--30},
  year={2020}
}

@inproceedings{
tramer2018slalom,
title={Slalom: Fast, Verifiable and Private Execution of Neural Networks in Trusted Hardware},
author={Florian Tramer and Dan Boneh},
booktitle={International Conference on Learning Representations},
year={2019},
url={https://openreview.net/forum?id=rJVorjCcKQ},
}

@inproceedings{liu2020client,
  title={Client-edge-cloud hierarchical federated learning},
  author={Liu, Lumin and Zhang, Jun and Song, SH and Letaief, Khaled B},
  booktitle={ICC 2020-2020 IEEE international conference on communications (ICC)},
  pages={1--6},
  year={2020},
  organization={IEEE}
}

@inproceedings{
farhat2025learning,
title={Learning to Specialize: Joint Gating-Expert Training for Adaptive MoEs in Decentralized Settings},
author={Yehya Farhat and Hamza ElMokhtar Shili and Fangshuo Liao and Chen Dun and Mirian Del Carmen Hipolito Garcia and Guoqing Zheng and Ahmed Hassan Awadallah and Robert Sim and Dimitrios Dimitriadis and Anastasios Kyrillidis},
booktitle={The Thirty-ninth Annual Conference on Neural Information Processing Systems},
year={2025},
url={https://openreview.net/forum?id=3FBByWp6GL}
}

@inproceedings{
zhou2025cryptomoe,
title={CryptoMoE: Privacy-Preserving and Scalable Mixture of Experts Inference via Balanced Expert Routing},
author={Yifan Zhou and Tianshi Xu and Jue Hong and Ye Wu and Meng Li},
booktitle={The Thirty-ninth Annual Conference on Neural Information Processing Systems},
year={2025},
url={https://openreview.net/forum?id=8pEqukyGrj}
}

@InProceedings{pmlr-v139-zhu21b,
  title = 	 {Data-Free Knowledge Distillation for Heterogeneous Federated Learning},
  author =       {Zhu, Zhuangdi and Hong, Junyuan and Zhou, Jiayu},
  booktitle = 	 {Proceedings of the 38th International Conference on Machine Learning},
  pages = 	 {12878--12889},
  year = 	 {2021},
  editor = 	 {Meila, Marina and Zhang, Tong},
  volume = 	 {139},
  series = 	 {Proceedings of Machine Learning Research},
  month = 	 {18--24 Jul},
  publisher =    {PMLR},
  url = 	 {https://proceedings.mlr.press/v139/zhu21b.html},

}

@article{zhan2021survey,
  title={A survey of incentive mechanism design for federated learning},
  author={Zhan, Yufeng and Zhang, Jie and Hong, Zicong and Wu, Leijie and Li, Peng and Guo, Song},
  journal={IEEE Transactions on Emerging Topics in Computing},
  volume={10},
  number={2},
  pages={1035--1044},
  year={2021},
  publisher={IEEE}
}

@article{liu2022gtg,
  title={Gtg-shapley: Efficient and accurate participant contribution evaluation in federated learning},
  author={Liu, Zelei and Chen, Yuanyuan and Yu, Han and Liu, Yang and Cui, Lizhen},
  journal={ACM Transactions on intelligent Systems and Technology (TIST)},
  volume={13},
  number={4},
  pages={1--21},
  year={2022},
  publisher={ACM New York, NY}
}

@article{borovac2022ensemble,
  title={Ensemble learning using individual neonatal data for seizure detection},
  author={Borovac, Ana and Gudmundsson, Steinn and Thorvardsson, Gardar and Moghadam, Saeed M and Nevalainen, P{\"a}ivi and Stevenson, Nathan and Vanhatalo, Sampsa and Runarsson, Thomas P},
  journal={IEEE journal of translational engineering in health and medicine},
  volume={10},
  pages={1--11},
  year={2022},
  publisher={IEEE}
}

@inproceedings{deencryptedllm,
  title={EncryptedLLM: Privacy-Preserving Large Language Model Inference via GPU-Accelerated Fully Homomorphic Encryption},
  author={de Castro, Leo and Escudero, Daniel and Agrawal, Adya and Polychroniadou, Antigoni and Veloso, Manuela},
  booktitle={Forty-second International Conference on Machine Learning}
}

@inproceedings{ICLR2025_6715b4e9,
 author = {Rho, Donghwan and Kim, Taeseong and Park, Minje and Kim, Jung Woo and Chae, Hyunsik and Ryu, Ernest and Cheon, Jung Hee},
 booktitle = {International Conference on Learning Representations},
 editor = {Y. Yue and A. Garg and N. Peng and F. Sha and R. Yu},
 pages = {41638--41664},
 title = {Encryption-Friendly LLM Architecture},
 url = {https://proceedings.iclr.cc/paper_files/paper/2025/file/6715b4e97be055687c1ecaf33913d358-Paper-Conference.pdf},
 volume = {2025},
 year = {2025}
}

@article{schick2023toolformer,
  title={Toolformer: Language models can teach themselves to use tools},
  author={Schick, Timo and Dwivedi-Yu, Jane and Dess{\`\i}, Roberto and Raileanu, Roberta and Lomeli, Maria and Hambro, Eric and Zettlemoyer, Luke and Cancedda, Nicola and Scialom, Thomas},
  journal={Advances in Neural Information Processing Systems},
  volume={36},
  pages={68539--68551},
  year={2023}
}

@inproceedings{yu2025survey,
  title={A survey on trustworthy llm agents: Threats and countermeasures},
  author={Yu, Miao and Meng, Fanci and Zhou, Xinyun and Wang, Shilong and Mao, Junyuan and Pan, Linsey and Chen, Tianlong and Wang, Kun and Li, Xinfeng and Zhang, Yongfeng and others},
  booktitle={Proceedings of the 31st ACM SIGKDD Conference on Knowledge Discovery and Data Mining V. 2},
  pages={6216--6226},
  year={2025}
}

@inproceedings{marsden2024universal,
  title={Universal test-time adaptation through weight ensembling, diversity weighting, and prior correction},
  author={Marsden, Robert A and D{\"o}bler, Mario and Yang, Bin},
  booktitle={Proceedings of the IEEE/CVF Winter Conference on Applications of Computer Vision},
  pages={2555--2565},
  year={2024}
}

\end{document}